\documentclass[journal,twoside,web]{ieeecolor}
\usepackage{xcolor,colortbl}
\usepackage{generic}
\usepackage{cite}
\usepackage{amsmath,amssymb,amsfonts}
\usepackage{algorithmic}
\usepackage{graphicx}
\usepackage{algorithm,algorithmic}
\usepackage{hyperref}
\usepackage{multirow}
\usepackage[caption=false]{subfig}
\usepackage{threeparttable}
\usepackage{orcidlink}
\hypersetup{colorlinks=true,
            linkcolor=blue,
            anchorcolor=blue,
            citecolor=blue}
\usepackage{textcomp}
\definecolor{mygreen}{RGB}{112,173,71}
\def\BibTeX{{\rm B\kern-.05em{\sc i\kern-.025em b}\kern-.08em
    T\kern-.1667em\lower.7ex\hbox{E}\kern-.125emX}}
\newcommand{\mvpa}{{{MVPA}}}
\newcommand{\cmke}{{{CMKE}}}
\newcommand{\mvketr}{{{MvKeTR}}}

\begin{document}
\title{MvKeTR: Chest CT Report Generation with Multi-View Perception and Knowledge Enhancement}
\author{Xiwei Deng\,\orcidlink{0009-0002-7682-4711}, Xianchun He\,\orcidlink{0009-0006-2909-8111}, Jianfeng Bao\,\orcidlink{0000-0002-8281-9401}, Yudan Zhou\,\orcidlink{0000-0002-3305-2072}, Shuhui Cai\,\orcidlink{0000-0003-2767-9490}, Congbo Cai\,\orcidlink{0000-0002-0600-8594}, and Zhong Chen\,\orcidlink{0000-0002-1473-2224}
\thanks{This work was supported by the National Natural Science Foundation of China (Grant No. 22161142024, 12375291, and 82071913). (Corresponding
authors: Zhong Chen; Congbo Cai.) }
\thanks{Xiwei Deng, Xianchun He, and Yudan Zhou are with the Institute of Artificial Intelligence, Xiamen University, Xiamen 361005, China (e-mail: xiweideng@stu.xmu.edu.cn; hexianchun@stu.xmu.edu.cn; ydzhou@stu.xmu.edu.cn).}
\thanks{Jianfeng Bao is with the Department of Magnetic Resonance Imaging, The First Affiliated Hospital of Zhengzhou University, Zhengzhou University, Zhengzhou, 450000, China}
\thanks{Shuhui Cai, Congbo Cai, and Zhong Chen are with the
Department of Electronic Science, Xiamen University, Xiamen 361005, China (e-mail: shcai@xmu.edu.cn; cbcai@xmu.edu.cn; chenz@xmu.edu.cn).}}

\maketitle
\begin{abstract}
CT report generation~(CTRG) aims to automatically generate diagnostic reports for 3D volumes, relieving clinicians' workload and improving patient care. Despite clinical value, existing works fail to effectively incorporate diagnostic information from multiple anatomical views and lack related clinical expertise essential for accurate and reliable diagnosis. To resolve these limitations, we propose a novel \textbf{M}ulti-\textbf{v}iew perception \textbf{K}nowledge-\textbf{e}nhanced \textbf{T}ansfo\textbf{R}mer~(MvKeTR) to mimic the diagnostic workflow of clinicians. Just as radiologists first examine CT scans from multiple planes, a Multi-View Perception Aggregator~(MVPA) with view-aware attention is proposed to synthesize diagnostic information from multiple anatomical views effectively. Then, inspired by how radiologists further refer to relevant clinical records to guide diagnostic decision-making, a Cross-Modal Knowledge Enhancer~(CMKE) is devised to retrieve the most similar reports based on the query volume to incorporate domain knowledge into the diagnosis procedure. Furthermore, instead of traditional MLPs, we employ Kolmogorov-Arnold Networks~(KANs) as the fundamental building blocks of both modules, which exhibit superior parameter efficiency and reduced spectral bias to better capture high-frequency components critical for CT interpretation while mitigating overfitting. Extensive experiments on the public CTRG-Chest-548K dataset demonstrate that our method outpaces prior state-of-the-art (SOTA) models across almost all metrics. The code is available at \href{https://github.com/xiweideng/MvKeTR}{https://github.com/xiweideng/MvKeTR}.
\end{abstract}

\begin{IEEEkeywords}
Radiology report generation, Multi-view learning, Knowledge enhancement, Kolmogorov-Arnold networks.
\end{IEEEkeywords}

\section{Introduction} 
\label{sec:intro}
\begin{figure}[!t]
\centering
\includegraphics[width=0.5\textwidth]{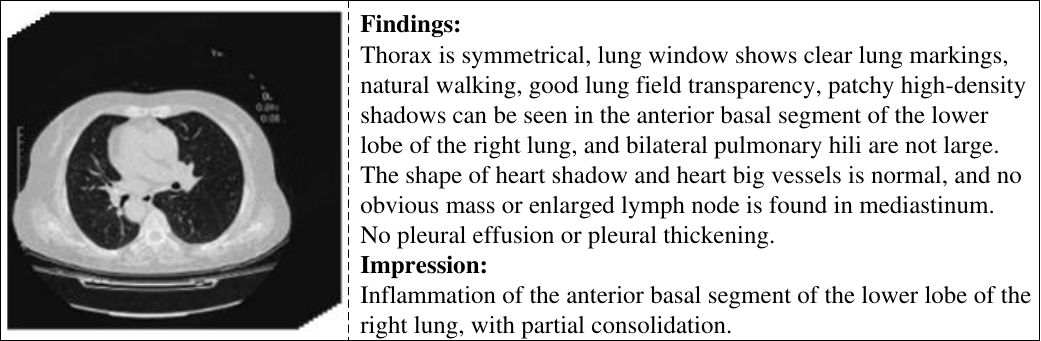}  
\caption{An example of chest CT volume and its corresponding report, including findings and impression.}  
\vspace{-1.5em}
\label{fig:example}    
\end{figure}

\IEEEPARstart{I}{n} clinical practice, medical imaging plays an indispensable role, serving as a cornerstone for disease diagnosis, report writing, and subsequent decision-making. After acquiring a patient's radiology images, physicians examine all anatomical structures and regions of concern, then employ related expert knowledge to write a hand-crafted, clinically coherent report that documents the observations~\cite{goergen2013evidence}. As evidenced in Fig.~\ref{fig:example}, such a radiology report typically
comprises a findings section that describes thorough medical observations, followed by an impression summarizing the most significant observations. The process of report writing is time-intensive and error-prone for clinicians~\cite{rosenkrantz2016us,rimmer2017radiologist,bruno2015understanding}.
Hence, automation of report generation holds substantial value in lightening physicians' burdens and improving the quality of reports. 

Despite recent advancements, there remain several constraints in previous studies that have not been fully addressed: (i) Ineffective incorporation of diagnostic information from multiple anatomical views. In comparison with 2D imaging~\cite{chen-etal-2021-cross-modal,jin2024improving}, 3D medical imaging (e.g., chest CT) provides a more holistic view of anatomical detail and spatial information, preserving the volumetric nature of the human body~\cite{muller2002computed}. 
Furthermore, unlike a single view of 3D volumes~\cite{tang2024work,hamamci2024ct2rep,chen2024dia}, the multi-view nature of CT scans enables physicians to examine the patient's condition from different anatomical views - axial, sagittal, and coronal- allowing for a more accurate diagnosis of many diseases. As a common example, pulmonary nodules, especially smaller ones, may be easily obscured by adjacent structures or misdiagnosed when viewed from a single perspective~\cite{setio2016pulmonary}. Axial views might reveal the presence of a nodule, but coronal and sagittal views are crucial for accurately assessing its characteristics and relationships to surrounding tissues~\cite{bankier2017recommendations}, thereby enhancing differential diagnosis~\cite{macmahon2017guidelines}. (ii) Absence of pertinent medical knowledge crucial for precise and dependable diagnosis. Relying merely on radiology images for report writing may be sub-optimal, overlooking the critical medical expertise~\cite{hamamci2024ct2rep,chen-emnlp-2020-r2gen}. Experienced clinicians usually consult analogous case reports to extract insights into diagnostic patterns and imaging features, guiding their decision-making when faced with complex or ambiguous cases during the diagnostic process. This knowledge-driven approach is critical for interpreting imaging findings, identifying subtle abnormalities, and writing clinically accurate reports.

Motivated by the aforementioned observations, we develop a novel Multi-view perception Knowledge-enhanced Transformer~(\mvketr) to overcome the deficiencies of current methods. In particular, to mirror how radiologists examine 3D CT volumes from multiple anatomical planes, we design a Multi-View Perception Aggregator (\mvpa) that utilizes view-aware attention to synthesize diagnostic information from different anatomical views effectively. To bridge the medical knowledge gap, we introduce a Cross-Modal Knowledge Enhancer (\cmke) that retrieves the most similar radiology reports, analogous to how physicians consult relevant clinical records. Additionally, we employ Kolmogorov-Arnold Networks (KANs)~\cite{liu2025kan}, which feature learnable nonlinear activation functions, to learn the complicated diagnostic relationships inherent in CT interpretation. Extensive experiments on the public CTRG-Chest-548K dataset showcase that our proposed~\mvketr~surpasses previous state-of-the-art methods across almost all established metrics.

In general, the main contributions of our work can be summarized as follows:
\begin{enumerate}
    \item We propose a novel Multi-view perception Knowledge-enhanced Transformer~(\mvketr) to achieve higher-quality CT report generation. To the best of our knowledge, \mvketr~is the first attempt to utilize multi-view diagnostic information, cross-modal medical expertise, and the potential of emerging KANs in this way.
    \item We recognize the inability of multi-view comprehension of current works and design a Multi-View Perception Aggregator that fuses diagnostic information from multiple anatomical views, capturing comprehensive spatial features effectively. 
    \item We introduce a Cross-Modal Knowledge Enhancer that leverages relevant reports to incorporate domain expertise into the report generation process, improving the clinical relevance and accuracy of the predicted reports. 
    \item Extensive comparison and ablation experiments showcase the superiority of our method over the recent state-of-the-art (SOTA) models.
\end{enumerate}

The remainder of this paper is structured as follows. Section~\ref{sec:related_works} reviews the related work. Section~\ref{sec:method} details our proposed model. Section~\ref{sec:exp} presents experimental results on a publicly available dataset. Section~\ref{sec:conclusion} concludes our work and points out crucial future work. 
\section{Related work}
\label{sec:related_works}
\subsection{Image Captioning}
Image captioning~\cite{bai2018survey,wang2020overview} aims to generate a brief human-like caption to describe an image. Traditional image captioning is template-based~\cite{yao2010i2t,socher2010connecting}, utilizing predefined templates with slots filled through property extraction and object detection. However, these methods suffer from rigid structural patterns, limiting their ability to generate diverse and natural descriptions. Neural network-based methods adopt an encoder-decoder architecture~\cite{vinyals2015show,rennie2017self,yao2010i2t}, which excels in capturing complex visual-linguistic relationships and generating more flexible and contextually appropriate captions. For instance, Xu et al.~\cite{xu2015show} created an attention-based model trained via deterministic back-propagation and stochastic maximization of a variational lower bound. Li et al.~\cite{li2017image} proposed a global-local attention (GLA) method by integrating local representation at the object-level with global representation at the image-level through an attention mechanism. Lu et al.~\cite{lu2017knowing} employed an adaptive attention model with a visual sentinel, enabling dynamic determination of whether and where to focus on the input images during sequential word generation. Cornia et al.~\cite{cornia2020meshed} introduced a meshed transformer with memory to learn a multi-level representation of the relationships between image regions and exploit low- and high-level features. Liu et al.~\cite{liu2021cptr} proposed CaPtion TransformeR (CPTR), a convolution-free architecture that takes serialized input images and instills global contextual awareness across encoder layers from the outset. These methods were primarily designed for natural scene images and may not directly apply to radiology images, necessitating lengthy reports with fine-grained medical observations.

\subsection{Radiology Report Generation}
\subsubsection{2D X-ray Report Generation}
The generation task of radiology reports can be regarded as the derivative task of image captioning, particularly for 2D X-rays. Motivated by the success of encoder-decoder architecture in image captioning, recent SOTA studies~\cite{jing2017automatic,xue2018mrma,hoogi2020natural,wang2021self} have all utilized this framework to produce medical reports combined with a visual extractor. Xue et al.~\cite{xue2018mrma} proposed a recurrent model combining Convolutional Neural Networks (CNNs) with Long Short-Term Memory (LSTM) networks, which integrates image encodings and previously generated sentences through attention guidance to ensure coherent report generation. Nooralahzadeh et al.~\cite{nooralahzadeh2021m2tr} introduced a curriculum learning-inspired consecutive generation framework that generates global concepts from medical images and then reforms them into detailed reports using a transformer-based architecture. Chen et al.~\cite{chen-etal-2021-cross-modal} designed a cross-modal memory network (CMN) that employs a shared memory to record image-text alignments and facilitate cross-modal interactions. To enhance global feature learning, Yi et al.~\cite{yi2024tsget} incorporated two-stage global enhancement layers into the transformer backbone. Specifically, the first layer is integrated to capture global visual context by establishing relationships between image-level global features and previously generated words, while the second layer is embedded to connect image-level global features with region-level information for comprehensive visual representations. Yi et al.~\cite{yi2024udt} presented the unsupervised disease tags (UDT) model, which obtains disease tag features via clustering and then fuses these with the attended visual features to assist report generation. To address performance imbalance between easy and difficult samples, Yi et al.~\cite{yi2024lhr} developed the Linear Hybrid-Reward based Reinforced Focal Learning (LHR-RFL). The framework, LHR-RFL, integrates a Linear Hybrid-Reward (LHR) module that quantifies learning difficulty via a linear weighting scheme combining BLEU-4, METEOR, and ROUGE-L metrics, and a Reinforced Focal Learning (RFL) mechanism that adaptively adjusts the contributions from difficult samples during training. Wang et al.~\cite{wang2024camanet} constructed a Class Activation Map-guided Attention Network (CAMANet), which enhances disease-pertinent feature representation by explicitly aligning aggregated class activation maps with attention mechanisms to strengthen cross-modal feature acquisition.
\subsubsection{3D CT Report Generation}
While these methods have demonstrated effectiveness in 2D X-ray imaging, the extension to 3D CT imaging has been underexplored, owing to the scarcity of publicly available paired image-report datasets~\cite{tang2024work,hamamci2024ct2rep} and the substantial computational resources~\cite{chen2024dia}. For Brain CT report generation, Zhang et al.~\cite{zhang2023weakly} designed WGAM-HI, a novel Weakly Guided Attention Model with Hierarchical Interaction. Specifically, a two-layer attention mechanism and report generator were employed to achieve many-to-many alignment between multiple CT images and report sentences. In addition, two weakly guided mechanisms based on pathological events and Grad-CAM were introduced to help the model focus on critical images and lesion areas, respectively. Furthermore, Zhang et al.~\cite{zhang2024co} developed CRHAN, a Co-occurrence Relationship Driven Hierarchical Attention Network, which incorporates two novel attention modules: a co-occurrence relationship guided semantic attention (CRSA) module to extract critical semantic features by embedding co-occurrence patterns of common pathologies, and a common-rare topic driven visual attention (CRVA) module to fuse common and rare semantic features for capturing important lesion information. This encoder-decoder architecture aims to improve the accuracy and diversity of generated reports by considering both common and rare pathological conditions. Regarding Chest CT report generation, Tang et al.~\cite{tang2024work} introduced SL-DG, which consists of a self-attention-based scan localizer (SL) for salient lesion feature extraction and a dynamic generator (DG) for report recommendation and synthesis. Hamamci et al.~\cite{hamamci2024ct2rep} established CT2rep, the first framework that generates chest CT reports using a novel auto-regressive causal transformer. By adapting LLaMA2-7B~\cite{touvron2023llama} for CT report generation, Chen et al.~\cite{chen2024dia} introduced Dia-llama to generate reports with diagnostic guidance prompts, which incorporates a disease-aware attention module and a disease prototype memory bank. To address the limitation of only considering global features, Chen et al.~\cite{chen2025large} proposed Reg2RG, a region-guided referring and grounding framework that leverages anatomical region masks and local feature decoupling to capture region-specific details with LLaMA2-7B~\cite{touvron2023llama} as the decoder. However, the aforementioned studies in 3D imaging 
often neglect the clinical value of multi-view diagnostic information inherent in volumetric medical imaging (e.g., chest CT scans). This oversight is critical because radiologists routinely analyze axial, coronal, and sagittal views to evaluate anatomical structures and pathological changes comprehensively. 
\subsubsection{Multi-view Strategies in Medical Image Analysis}
Multi-view strategies have been applied to medical image segmentation (e.g., liver segmentation~\cite{xia2024automatic}, lung tumor segmentation~\cite{liu2024multi}) and 2D X-ray report generation (e.g., chest X-rays~\cite{yuan2019automatic,yang2020automatic}). Yet, the application in 3D CT report generation remains underexplored. In medical image segmentation, Xia et al.~\cite{xia2024automatic} introduced a hybrid framework combining 2D Dual Self-Attention (DSA) networks on multiplanar slices with lightweight 3D fusion and full connection condition random field refinement, achieving Dice scores exceeding 95\% by resolving ambiguities in low-context regions. Similarly, Liu et al.~\cite{liu2024multi} proposed a multi-scale multi-view network(MSMV-Net) for lung tumor segmentation, integrating multi-scale features from axial, coronal, and sagittal views via attention-weighted fusion to address feature incompatibility in 3D U-Net architectures. These works highlight the efficacy of multi-view fusion in pixel-level localization but lack higher-level semantic reasoning capabilities (e.g., linking nodule characteristics to clinical findings) required for diagnostic report generation. In 2D X-ray report generation, Yuan et al.~\cite{yuan2019automatic} pioneered cross-view consistency enforcement between frontal and lateral X-rays through a generative encoder-decoder framework, synthesizing multi-view visual features via a sentence-level attention mechanism in a late fusion fashion to preserve complementary information across views. Yang et al.~\cite{yang2020automatic} introduced multi-modal attention to align visual features with medical concepts. However, these approaches are constrained by the simplicity of 2D planar X-ray imaging and short descriptions, failing to address the volumetric complexity and nuanced diagnostic requirements of 3D CT imaging.
\subsubsection{Professional Knowledge in Report Generation}
Recent attempts~\cite{yang2023m2kt,kale2023knowledge,hou2023mkcl,huang2023kiut} have explored diverse strategies to integrate medical expertise into radiology report generation, which can be broadly categorized into knowledge base-driven and knowledge graph-based frameworks. While these methods demonstrate progress, there are several limitations. Concerning knowledge base-driven approaches, Yang et al.~\cite{yang2023m2kt} developed a report generation model that leverages a learnable medical knowledge base and multi-modal alignment mechanism to learn visual features comprehensively. However, this approach suffers from limited interpretability of the diagnostic process through the dynamic knowledge update mechanism. For knowledge graph-based models, Kale et al.~\cite{kale2023knowledge} proposed KG-BART, which relies on knowledge graphs of ten abdominal organs and a prepared radiology dictionary. Hou et al.~\cite{hou2023mkcl} proposed a Medical Knowledge with Contrastive Learning model (MKCL), which employs a chest abnormality graph to enhance lesion detection via contrastive learning. Huang et al.~\cite{huang2023kiut} developed a Knowledge-injected U-Transformer (KiUT), introducing symptom graphs to inject clinical knowledge into generation. Despite achieving improvements, these methods require labor-intensive pre-construction of domain-specific graphs.

\begin{figure*}[!t]
\centering
\includegraphics[width=0.9\textwidth]{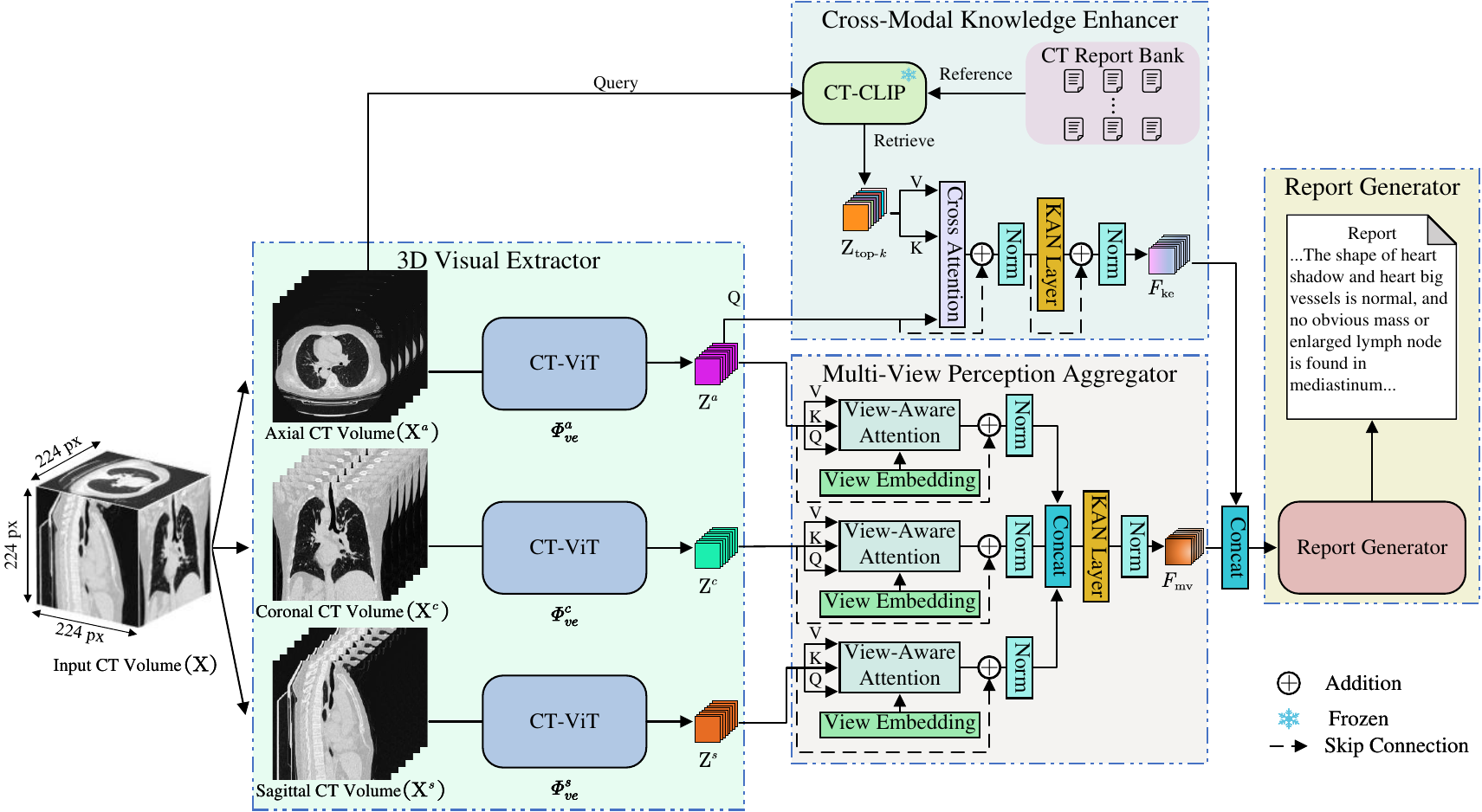}  
\caption{The overview of our proposed \mvketr,~which can be partitioned into four parts: 3D visual extractor, multi-view perception aggregator, cross-modal knowledge enhancer, and report generator. The 3D visual extractor extracts CT patches from the axial, coronal, and sagittal views of the input 3D CT volume. The multi-view perception aggregator aggregates diagnostic information from multiple anatomical views effectively. The cross-modal knowledge enhancer incorporates relevant clinical expertise into the diagnosis pipeline. The report generator predicts the final reports relying on the multi-view diagnostic information and the related professional knowledge.}  
\label{fig:architecture}    
\end{figure*}

\begin{figure*}[!t]
\centering
\includegraphics[width=0.9\textwidth]{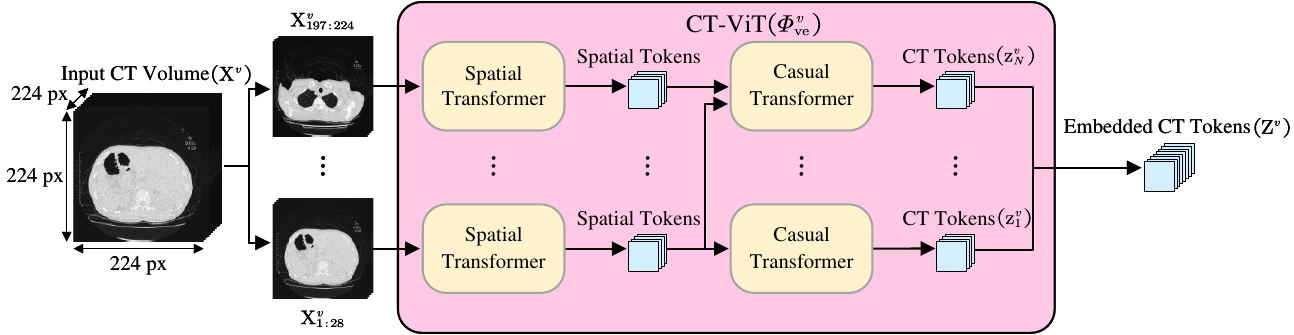}  
\caption{Illustration of vision feature extraction pipeline through CT-ViT.}  
\label{fig:pipeline}    
\end{figure*}
\section{Methodology}
\label{sec:method}
In this section, we elaborate on our proposed \mvketr. Table~\ref{table:notations} illustrates the key notations and their respective descriptions adopted in this study.

\begin{table}[!t]
\caption{Key Notations}
\setlength{\tabcolsep}{1mm}
\centering
\begin{tabular}{ll}
    \hline
    Notations & Descriptions \\
    \hline
   $\mathrm{X}$&The input 3D CT volume.\\
   $\mathrm{N}_{p}$&The number of extracted CT patches.\\
   $\mathrm{N}_{r}$&The number of reports in the CT report bank.\\
   $\varPhi_{ve}^a$, $\varPhi_{ve}^c$, $\varPhi_{ve}^s$ & The axial, coronal, and sagittal CT-ViT.\\
    $\mathrm{X}^a$, $\mathrm{X}^c$, $\mathrm{X}^s$ & The axial, coronal, and sagittal 3D CT volume.\\
    $\mathrm{Z}^a$, $\mathrm{Z}^c$, $\mathrm{Z}^s$ & The axial, coronal, and sagittal embedded CT tokens.\\
    $B$&The batch size.\\
    $D$&The dimension of the embedding.\\
    $h,w$&The height and width of the CT slices\\
    $P_t$& The temporal patch size.\\
    $P_h, P_w$ & The spatial patch sizes of height and width. \\
    $N_t$& The number of temporal patches.\\
    $\mathrm{Z}_{\text{top-$k$}}$&The retrieved top-$k$ report embeddings.\\
    $F_{\mathrm{ke}}$&The features encoded by cross-modal knowledge enhancer.\\
    $F_{\mathrm{mv}}$&The features encoded by multi-view perception aggregator.\\
    $\gamma$&The model parameters.\\
    $\Phi_k$&Nonlinear learnable parameters of the $k^{th}$ KAN Layer.\\
    $E_v$&The view embedding of anatomical view $v$.\\
    \hline
\end{tabular}
\label{table:notations}
\end{table}

\subsection{Overview of the Proposed Approach}
The generation of radiology reports can be considered an image-to-text problem, for which we follow a sequence-to-sequence paradigm. In doing so, unlike previous approaches~\cite{tang2024work,chen2024dia,hamamci2024ct2rep} that typically process a single-view image, our network handles the input 3D CT volume $\mathrm{X}\in{\mathbb{R}^{(224)\times 224\times 224}}$ by transforming it into three source sequences of CT patches: $\mathrm{Z}^v =\{\mathrm{z}^v_{1}, \mathrm{z}^v_{2}, ..., \mathrm{z}^v_{N}\}, \mathrm{z}^v_{n}\in \mathbb{R}^{(8)\times 8 \times 8}$,
where $v \in \{a, c, s\}$ represents the axial, coronal, and sagittal views respectively, $\mathrm{Z}^v$ are embedded CT tokens extracted by 3D visual extractor for each view, and $D$ is the dimension of the embedding. We regard the corresponding report as the target sequence $Y=\{y_{1}, y_{2}, ..., y_{L}\}$, where $y_{l}\in \mathbb{V}$ are the predicted tokens, $L$ the length of predicted tokens and $\mathbb{V}$ the vocabulary of all possible tokens. Thus, the aforementioned process of report generation can be formalized as follows: 
\begin{equation}
\log p(Y|X)=\sum_{l=1}^T \log p\left(y_l | y_1, \ldots, y_{l-1}, X\right)
\end{equation}
Subsequently, the model is trained to maximize $p(Y|X)$ by the negative conditional log-likelihood of  $Y$ given $X$: 
\begin{equation}
\gamma^*=\underset{\gamma}{\arg \max } \sum_{l=1}^T \log p\left(y_l | y_1, \ldots, y_{l-1}, X; \gamma\right)
\end{equation}
where $\gamma$ is the parameters of the model. The overall architecture of our proposed framework is illustrated in Fig.~\ref{fig:architecture}, where the contributions of each component are detailed in the following subsections.
\subsection{3D Visual Extractor}
Following the previous work~\cite{hamamci2024ct2rep}, CT-ViT is employed to extract embedded CT tokens $\mathrm{Z}^v\in{\mathbb{R}^{8\times 8\times 8 \times 512}}$, by initially extracting \( (28) \times 28 \times 28\) non-overlapping patches from the given 3D CT volume $\mathrm{X}^v$, where $v \in \{a, c, s\}$. Then, each patch is embedded in a D-dimensional space, reshaped, and linearly transformed to an intermediate tensor \( T_i \in \mathbb{R}^{B \times N_t \times \frac{h}{P_h} \times \frac{w}{P_w} \times D}\). Here, \( P_t \) represents the temporal patch size, \( B \) the batch size, \( N_t \) the temporal patch count, \( h \) and \( w \) correspond to the height and width of the CT slices, and $D$ the dimension of the embedding space. \( P_h \) and \( P_w \) denote the spatial patch sizes of height and width. Subsequently, $T_i$ is processed and reshaped to the final outcome tensor \( T_o \in \mathbb{R}^{\left( \frac{h}{P_h} \cdot \frac{w}{P_w} \right) \times \left( B \cdot N_t \right) \times D}\) consecutively by the spatial and causal transformer models. The 3D visual extractor comprises three independent CT-ViTs, which process the axial, coronal, and sagittal view CT volume, respectively. Fig.~\ref{fig:pipeline} illustrates the pipeline of visual feature extraction through CT-ViT.

The whole procedure can be formally formulated as:
\begin{align}
    &\mathrm{Z}^a =\{\mathrm{z}^a_{1}, \mathrm{z}^a_{2}, ..., \mathrm{z}^a_{N_p}\} = \varPhi_{ve}^a(\mathrm{X^a})\\
     &\mathrm{Z}^c =\{\mathrm{z}^c_{1}, \mathrm{z}^c_{2}, ..., \mathrm{z}^c_{N_p}\} = \varPhi_{ve}^c(\mathrm{X^c})\\
          &\mathrm{Z}^s =\{\mathrm{z}^s_{1}, \mathrm{z}^s_{2}, ..., \mathrm{z}^s_{N_p}\} =\varPhi_{ve}^s(\mathrm{X^s})
\end{align}
where $\mathrm{N}_{p}$ denotes the number of extracted CT patches, $\varPhi_{ve}^a(\cdot)$, $\varPhi_{ve}^c(\cdot)$, and $\varPhi_{ve}^s(\cdot)$ denote the axial, coronal, and sagittal CT-ViT, respectively.
\vskip 0.5em
\subsection{KAN as Effective Successor}
Multi-layer perceptrons (MLPs)~\cite{rumelhart1986learning,hornik1989multilayer} serve as basic blocks of contemporary deep learning models and prevail in numerous tasks (e.g., radiology report generation). However, two critical limitations emerge when applying typical MLPs to CT report generation: quadratic parameter scaling with network depth (Theorem 3.3~\cite{wang2025on}) and spectral bias favoring low-frequency patterns~\cite{rahaman2019spectral}. These limitations render MLPs susceptible to overfitting when learning high-frequency components (e.g., ground-glass opacities, micronodules) from multi-view CT imaging data.

Inspired by the superior parameter efficiency and reduced spectral bias of the Kolmogorov-Arnold Network (KANs)~\cite{liu2025kan}, we explore using KANs as an alternative to traditional MLPs to address the above limitations. First, Theorem 3.2~\cite{wang2025on} establishes that any MLP with the $\mathrm{ReLU}^k$ activation function can be represented as a KAN with a comparable size, which shows KANs can achieve at least equivalent expressiveness as MLPs. Second, Theorem 3.3~\cite{wang2025on} proves that MLP parameterization scales quadratically with grid size ($O(G^2W^4L)$), whereas KANs maintain linear parameter scaling ($O(GW^2L)$), where G is the number of spline interpolation points, W the number of neurons per hidden layer, and L the number of layers. This demonstrates that KANs achieve comparable expressiveness with significantly fewer parameters. For CT report generation, where fine-grained details (e.g., subtle lesions) require capturing intricate nonlinear patterns, KANs avoid the parameter bloat of MLPs, inherently reducing the risk of overfitting.

Conventional MLPs employing ReLU activations (or even tanh) are documented to exhibit spectral bias~\cite{rahaman2019spectral}, manifesting as a preferential fitting of low-frequency components. In the subsequent analysis, we investigate the spectral bias of KANs. As proven in Theorem 4.1(~\cite{wang2025on}), KANs exhibit less spectral bias than MLPs. Specifically, the eigenvalue ratio of the Hessian matrix satisfies: 
\begin{equation}
    \frac{\lambda_{N}(M)}{\lambda_{d'(d-1) + 1}(M)} \leq Cd
\end{equation}
for a constant $C$ depending only on the spline degree $k$. This property enables KANs to learn high-frequency components (edges, textures, anomalies) more effectively, which is critical for capturing subtle pathological patterns in CT scans.

\begin{figure}[!t]
\centering
\includegraphics[width=0.35\textwidth]{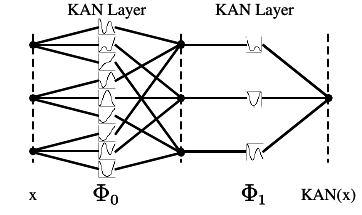}  
\caption{The architecture of Kolmogorov–Arnold Networks (KANs) with a series of KAN layers.}  
\label{fig:kan}    
\end{figure}

Now, we detail the architecture of KANs~\cite{liu2025kan}. Given an input $\mathbf{x}$, a KAN network is a stack of multiple KAN layers(as illustrated in Fig.~\ref{fig:kan}), which can be characterized
as:
\begin{equation}
    \operatorname{KAN}(\mathbf{x})=\left(\boldsymbol{\Phi}_{k-1} \circ \boldsymbol{\Phi}_{k-2} \circ \cdots \circ \boldsymbol{\Phi}_{1} \circ \boldsymbol{\Phi}_{0}\right) \mathbf{x}
\end{equation}
The learnable activation functions $\phi$ of the $k^{th}$ KAN layer can be formalized as
\begin{equation}
    \boldsymbol{\Phi}_{k}=\{\phi_{i,j}\},\quad i=1,2,\cdots,N_{\mathrm{in}},\quad j=1,2\cdots,N_{\mathrm{out}}
\end{equation}
where $\phi_{i,j}$ represents functions with trainable parameters, and $N_\mathrm{in}$ and $N_\mathrm{out}$ denote the input and output dimensions, respectively.
Then, the calculation of the KAN network from the $k^{th}$ layer to the $(k+1)^{th}$ layer can be represented as:
\begin{equation}
\mathbf{x}_{k+1} = \boldsymbol{\Phi}_{k}\mathbf{x}_{k}
\end{equation}
Here, $\boldsymbol{\Phi}_{k}$ is defined as a matrix form:
\begin{equation}
\boldsymbol{\Phi}_{k} = \left(\begin{array}{cccc}
\phi_{k, 1,1}(\cdot) & \phi_{k, 1,2}(\cdot) & \cdots & \phi_{k, 1, n_{k}}(\cdot) \\
\phi_{k, 2,1}(\cdot) & \phi_{k, 2,2}(\cdot) & \cdots & \phi_{k, 2, n_{k}}(\cdot) \\
\vdots & \vdots & & \vdots \\
\phi_{k, n_{k+1}, 1}(\cdot) & \phi_{k, n_{k+1}, 2}(\cdot) & \cdots & \phi_{k, n_{k+1}, n_{k}}(\cdot)
\end{array}\right)
\end{equation}

\begin{figure}[!t]
\centering
\label{fig:diagram}
\subfloat[Attention]{
    \label{fig:attention}
    \includegraphics[scale=0.8]{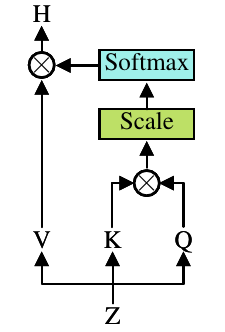}}
    \hfill
\subfloat[View-aware Attention]{
    \label{fig:view-aware attention}
    \includegraphics[scale=0.8]{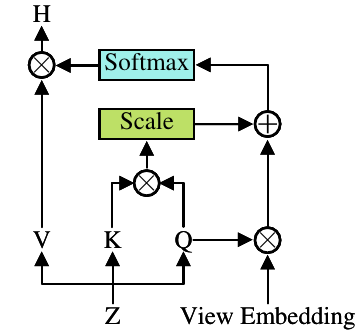}}
\caption{The schematic diagram of Attention and View-aware Attention. \text{``}$\otimes$\text{''} denotes matrix multiplication. \text{``}$\oplus$\text{''} represents element-wise addition.}
\end{figure}

\subsection{Multi-View Perception Aggregator}
The multi-view perception aggregator is devised to perceive and aggregate diagnostic information from three orthogonal planes of the input CT volume: axial, coronal, and sagittal views. The details are presented below.
\subsubsection{View-aware Attention}
The vanilla attention(see Fig.~\ref{fig:attention}) adopted in Transformer\cite{Vaswani2017atten} is computed through the correlation between the query matrix $Q$ and the key matrix $K$, which is then used to aggregate information from the value matrix $V$. The process can be formulated as follows:
\begin{equation}
\operatorname{Attention}(Q,K,V) = \operatorname{Softmax}\left(\frac{QK^T}{\sqrt{d_k}}\right)V
\end{equation}
where $Q = ZW_q$, $K = ZW_k$, $V = ZW_v$ are projected from the same input $Z$, $W_q \in \mathbb{R}^{D \times d_k}$, $W_k \in \mathbb{R}^{D \times d_k}$, $W_v \in \mathbb{R}^{D \times d_v}$ are learnable weight matrices. 

This mechanism has proven effective in various tasks, particularly in capturing global dependencies. However, it fails to perceive the inherent distinctiveness between different views of the input CT volume, which can lead to suboptimal performance.

To effectively model the view-specific patterns, we design a view-aware attention module(see Fig.~\ref{fig:view-aware attention}), which introduces view embeddings $E_v$ to characterize the uniqueness of each anatomical view explicitly. To be specific, given the query $Q$, key $K$, value $V$ matrices, as well as $E_v$, view-aware attention is formulated as:
\begin{equation}\label{eq:view_aware_attention}
\begin{aligned}
\mathrm{VAA_v} &= \operatorname{ViewAwareAttention}(Q, K, V, E_v)\\
&=\operatorname{Softmax}\left(\frac{QK^T}{\sqrt{d_k}}+QE_v^T\right)V
\end{aligned}
\end{equation}
where $E_v \in \mathbb{R}^{N \times d_k}$, $v \in \{a, s, c\}$.

\subsubsection{Multi-view Aggregation}
After obtaining view-specific representations through Eq.~\eqref{eq:view_aware_attention}, we propose a multi-view aggregation strategy to effectively combine the complementary information from multiple anatomical views while preserving their distinctive characteristics. First, for each view $v$, we normalize the attended values with residual connections:
\begin{equation}
\mathrm{AN_v} = \mathrm{Norm}(\mathrm{VAA_v} + Z^v)
\end{equation}
where $Z^v$ denotes the embedded CT tokens from view $v$. 

To aggregate information across different views, we concatenate the normalized outputs from the axial, sagittal, and coronal views. This concatenated representation is then fused by a KAN layer, followed by a final normalization:
\begin{equation}
\begin{aligned}
\mathrm{Concat}_{mv}&=[\mathrm{AN_a}; \mathrm{AN_s}; \mathrm{AN_c}] \\
F_{\text{mv}} &= \mathrm{Norm}(\mathrm{KANLayer}(\mathrm{Concat}_{mv}))
\end{aligned}
\end{equation}
Here, $[\cdot;\cdot;\cdot]$ denotes the concatenation operation. We utilize the KAN instead of the traditional MLP layer to capture complex interactions between different views.

\begin{figure}[!t]
\centering
\includegraphics[width=0.4\textwidth]{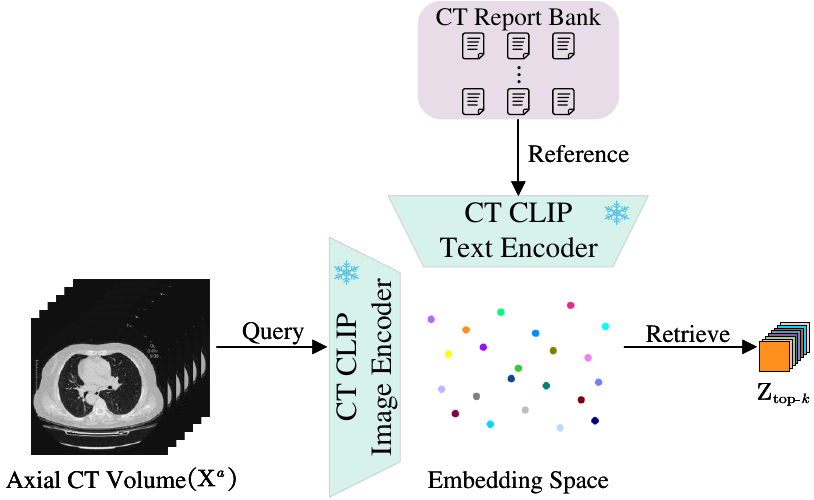}  
\caption{Illustration of retrieving top-$k$ report embeddings given a query CT volume.}  
\label{fig:retrieve}    
\end{figure}

\subsection{Cross-Modal Knowledge Enhancer}
Analogous to clinicians' diagnostic practice of consulting relevant medical records, the cross-modal knowledge enhancer is developed to integrate medical expertise into report generation through a multi-stage adaptive fusion procedure, designed to balance prior knowledge and novel observations. The entire procedure comprises two phases, as detailed below.
\subsubsection{Volume-to-Report Retrieval}
As depicted in Fig.~\ref{fig:retrieve}, the cross-modal knowledge enhancer performs volume-to-report retrieval through a well-trained CT-CLIP model on the CT-RATE~\cite{hamamci2024ct-clip}. Specifically, CT-CLIP is a contrastive report-volume model consisting of an image encoder for axial 3D chest CT volumes and a text encoder for radiology reports. To our knowledge, CT-RATE currently stands as the largest open-source dataset pairing 3D chest CT volumes with radiology reports, comprising 25,692 non-contrast CT volumes (expanded to 50,188 through reconstructions) from 21,304 patients. This dataset ensures clinical reliability through three key characteristics: (1) Comprehensive coverage of diverse pathologies, demographic profiles (age/sex), and imaging parameters (slice thickness/manufacturers); (2) Rigorous quality control with board-certified radiologist validation of all reports; (3) Robust abnormality labeling via RadBERT-based automated classifier demonstrating F1 scores exceeding 0.85 across 18 pathological categories~\cite{yan2022radbert}, manually verified across 1,000 reports. This large-scale, meticulously curated dataset minimizes the risk of retrieving erroneous or biased reports, as the retrieval process inherently prioritizes clinically validated cases through CT-CLIP’s semantic alignment in the shared embedding space. Given an input axial CT volume $\mathrm{X}^a$ and a CT report bank $\mathrm{R} = \{R_1, R_2, ..., R_{N_r}\}$ where $R_i$ represents the i-th radiology report, $N_r$ is the number of reports, we aim to retrieve the top-$k$ most relevant report embeddings $\mathrm{Z}_{\text{top-$k$}}$ from $\mathrm{R}$. Here, to maintain consistency with CT-CLIP's self-supervised pre-training, we utilize axial CT volume as query images for report retrieval.

The CT-CLIP image encoder $\mathrm{E}_\mathrm{image}$ maps the input volume $\mathrm{X}^a$ to a D-dimensional embedding $v\in{\mathbb{R}^{D}}$:
\begin{equation}
v = \operatorname{E}_{\text{image}}(\mathrm{X}^a)
\end{equation}

\begin{figure}[!t]
\centering
\includegraphics[scale=0.8]{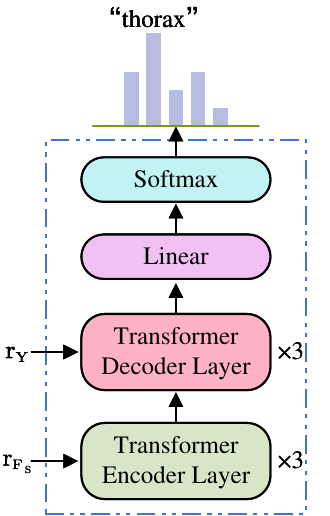}  
\caption{Illustration of encoder-decoder architecture in report generator.}  
\label{fig:rg}    
\end{figure}

Similarly, each report $R_i$ is encoded into a report embedding $r_i\in{\mathbb{R}^{D}}$ by the CT-CLIP text encoder $\mathrm{E}_\mathrm{text}$:
\begin{equation}
r_i = \operatorname{E}_{\text{text}}(R_i)
\end{equation}

After encoding both the input CT volume and reports into the shared embedding space, we apply L2 normalization to standardize the embeddings and compute the cosine similarity between them to retrieve the most relevant report embeddings:
\begin{align}
\operatorname{sim}(v,r_i) &= \frac{v^T r_i}{||v|| ||r_i||} \\
\operatorname{Z}_{\text{top-}k} &= \underset{r_i \in \mathrm{R'} }{\arg \max}^{\text{top-}k} \operatorname{sim}(v,r_i)
\end{align}
where $\mathrm{R'}=\{r_1, r_2, ..., r_{N_r}\}$ is the set of report embeddings.
\subsubsection{Knowledge Enhancement}

To effectively integrate visual features with retrieved medical knowledge, we introduce a knowledge enhancement mechanism. This mechanism employs a cross-attention module that enables the model to adaptively focus on relevant information from both the axial embedded CT tokens $Z^a$ and the retrieved report embeddings $Z_{\text{top-k}}$. Specifically, the cross-attention mechanism assigns learnable weights to the retrieved reports via scaled dot-product operations, prioritizing features consistent with the visual evidence from the input CT volume while suppressing conflicting or low-confidence information. For example, if retrieved reports contain contradictory findings (e.g., conflicting descriptions of nodule characteristics), the attention weights inherently downweight such inconsistencies, ensuring the model's reliance on both visual evidence and clinically validated knowledge. Formally, given the query $Q$, key $K$, value $V$ matrices, the cross-attention can be  computed as follows:
\begin{equation}
\begin{aligned}
\operatorname{CA_a} &= \text{CrossAttention}(Q,K,V) \\
&= \operatorname{Softmax}\left(\frac{QK^T}{\sqrt{d_k}}\right)V
\end{aligned}
\end{equation}
where $Q = Z^aW_q$, $K = Z_{\text{top-k}}W_k$, $V = Z_{\text{top-k}}W_v$, $W_q$, $W_k$, and $W_v$ are learnable projection matrices. The output is further processed through residual connection and layer normalization:
\begin{equation}
\operatorname{AN_a} = \text{Norm}(\mathrm{CA_a} + Z^a)
\end{equation}
Here, the residual connection preserves the original visual features as direct observations. This prevents retrieved knowledge (prior) from overriding novel findings, ensuring the model retains the capacity to detect unseen abnormalities.
Then, we process the enhanced features through a KAN layer, followed by another residual connection and layer normalization:
\begin{equation}
F_{\text{ke}} = \text{Norm}(\text{KANLayer}(\mathrm{AN_a}) + \mathrm{AN_a})
\end{equation}

\subsection{Report Generator}
To efficiently produce reports, a significant body of recent works~\cite{nooralahzadeh2021m2tr, chen-etal-2021-cross-modal, yang2023m2kt, yi2024tsget, yi2024udt, wang2024camanet, hamamci2024ct2rep, tang2024work, chen2024dia} adopt the encoder-decoder architecture, which is built upon the standard Transformer. We follow the R2GenCMN proposed by Chen et al.~\cite{chen-etal-2021-cross-modal}, which enhances the alignment between visual and textual modalities by introducing cross-modal memory networks(CMN). In detail, this network employs a learnable memory matrix and revises memory vectors by the attention mechanism during the multi-threaded querying and responding procedure, so as to better align the visual and textual representations. For an input 3D CT volume, We first concatenate the multi-view diagnostic features $F_{\mathrm{mv}}$ and knowledge-enhanced features $ F_{\mathrm{ke}}$ to obtain the fused representation $F_S=\left[F_{\mathrm{mv}};F_{\mathrm{ke}}\right]$, which implements a Bayesian posterior estimation:
\begin{equation}
p(Y \mid X, K) \propto \underbrace{p(X \mid Y)}_{\text {Likelihood }\left(F_{\text {mv }}\right)} \cdot \underbrace{p(K \mid Y)}_{\text {Prior }\left(F_{\text {ke }}\right)}  
\end{equation}
where $F_{\mathrm{mv}}$ denotes direct CT observations (e.g., ground-glass opacities) as the likelihood term, and $F_{\mathrm{ke}}$ provides evidence-based priors from retrieved reports (e.g., typical nodule descriptions). The report generator then utilizes the fused representation $F_S$ to generate the final report, emulating radiologists' diagnostic workflow of combining evidence and domain knowledge. Here, retrieved knowledge complements direct diagnostic observations instead of replacing them, ensuring the integration of prior domain expertise. By doing so, our model retains its ability to identify novel clinical findings not present in retrieved cases. Given a source sequence $F_S=\left\{f_1, f_2, f_3, \ldots, f_S\right\}$, a target sequence $Y=\left\{y_1, y_2, y_3, \ldots, y_T\right\}$, and a memory matrix $M=\left\{m_1, m_2, m_3, \ldots, m_l\right\}$, the memory responses of $F_S$ and $Y$ can be obtained by: 
\begin{align}
r_{F_S} & =\frac{F_S \cdot m_l^T}{\sqrt{d}} \cdot m_l \\
r_{y_t} & =\frac{Y_T \cdot m_l^T}{\sqrt{d}} \cdot m_l
\end{align}
where $l$ denotes the number of memory vectors, $\mathbf{m}_j \in \mathbb{R}^d \text { the memory vector at row } j$ with $d$ indicating its dimension. As illustrated in Fig.~\ref{fig:rg}, the report generator of our model integrates a three-layer transformer with the above-mentioned cross-modal memory mechanism. 
The features encoded by \mvpa~and \cmke~modules are concatenated and fed into this module, where the encoder processes only multi-modal features while the decoder takes only textual features as input. This process can be formulated as:
\begin{align}
Z_S& =R_e(r_{f_1}, r_{f_2}, \ldots, r_{f_S})\\
y_t& =R_d(Z_S, r_{y_1}, r_{y_2}, \ldots, r_{y_{(t-1)}})
\end{align}
where $R_d(\cdot)$ and $R_e(\cdot)$ denote the decoder and encoder of the report generator, respectively, $\left\{r_{f_1}, r_{f_2}, \ldots, r_{f_S}\right\}$ and $\left\{r_{y1}, r_{y2}, \ldots, r_{y(t-1)}\right\}$ represent the memory responses for multi-modal features and textual features from previously generated tokens, respectively, $Z_S$ denotes the intermediate encoded states, $y_t$ the predicted output at the current time step $t$. The complete report is produced by:
\begin{equation}
    Y_t=\operatorname{Softmax}\left(\operatorname{Linear}\left(y_t\right)\right)
\end{equation}
where $\operatorname{Linear}(\cdot)$ and $\operatorname{Softmax}(\cdot)$ denote the Linear and Softmax layer of the module, respectively, and $Y_t$ represents the predicted token at time step $t$.

\begin{figure}[!t]
\centering
\includegraphics[width=0.4\textwidth]{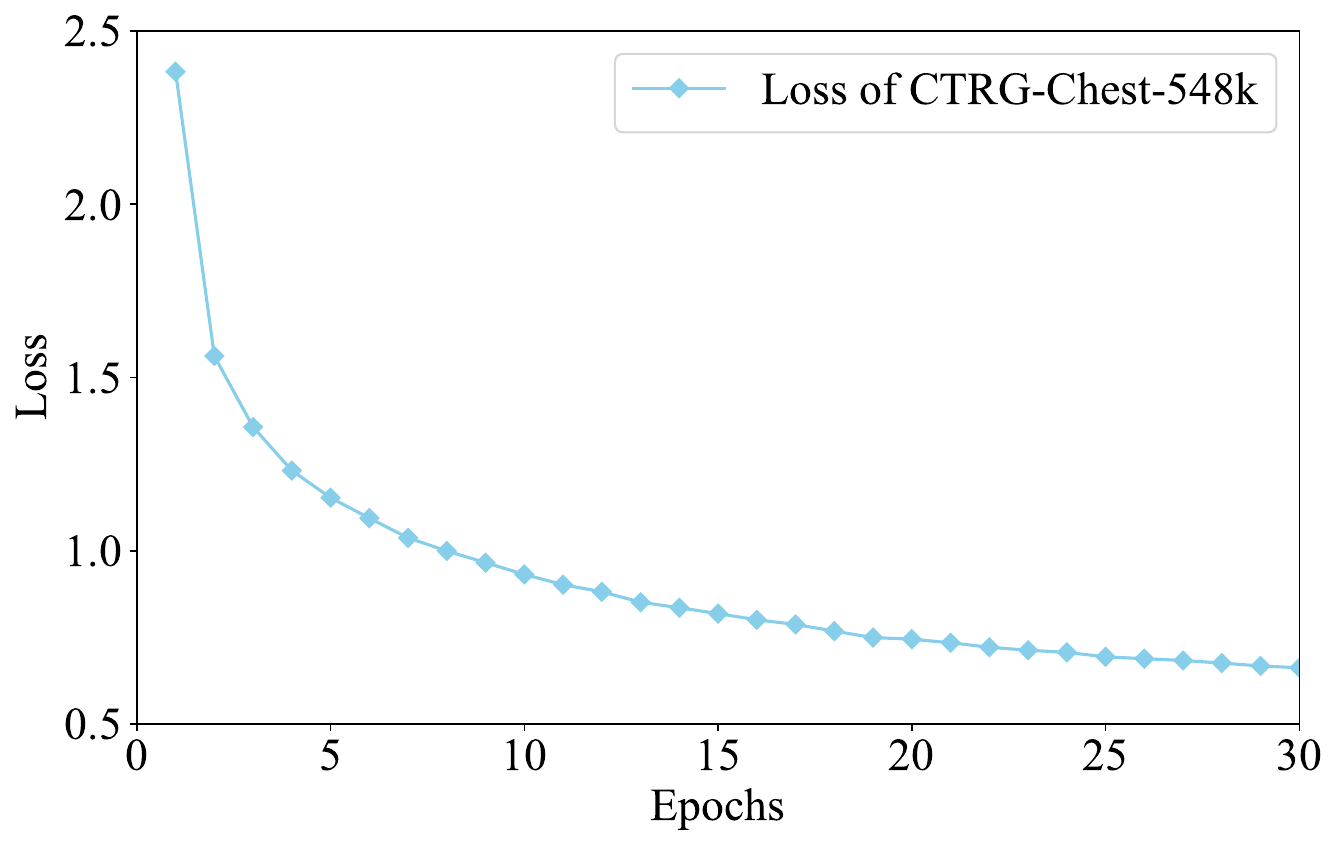}  
\caption{The loss curve during training on the CTRG-Chest-548K.}  
\label{fig:loss_curve}    
\end{figure}

\section{Experiments}
\label{sec:exp}
\subsection{Dataset}
To validate the superiority of our proposed model, extensive experiments are carried out on the public chest CT report dataset CTRG-Chest-548K~\cite{tang2024work}\footnote{\url{https://github.com/tangyuhao2016/CTRG}}, which encompasses 1,804 image-report pairs. Conforming to the same split proportion~\cite{tang2024work}, 80\% of the dataset is randomly allocated for training and 20\% for testing. 

As evident in Fig.~\ref{fig:loss_curve}, we present the loss value of the model trained on the CTRG-Chest-548K dataset. We can observe that the loss curve exhibits a consistent downward trend and eventually converges to a stable state.

\subsection{Evaluation Metrics}
The quality of the predicted reports is accessed through the most widely used evaluation metrics\footnote{\url{https://github.com/tylin/coco-caption}}: BLEU (i.e., BLEU-1,
BLEU-2, BLEU-3, and BLEU-4)~\cite{papineni-etal-2002-bleu}, METEOR~\cite{denkowski2011meteor}, and ROUGE-L~\cite{lin-2004-rouge}. BLEU (Bilingual Evaluation Understudy) computes the overlap of n-grams between the candidate and reference reports and incorporates a brevity penalty to discourage overly short candidate sentences. METEOR (Metric for Evaluation of Translation with Explicit ORdering) extends beyond standard exact word matching by incorporating stemming and synonymy matching, which builds on the harmonic mean of unigram precision and recall, with more emphasis on recall than precision. ROUGE-L(Recall-Oriented Understudy for Gisting Evaluation-Longest Common Subsequence) calculates the longest common subsequence between the candidate and reference reports, allowing non-consecutive matches while maintaining word order.
\subsection{Implementation Details}
For both the compared methods and our method, CT-ViT~\cite{hamamci2024generatect} is employed as the 3D visual extractor, while all other configurations for the compared methods remain consistent with the original paper. The resolution of each CT volume is resized to \(224 \times 224 \times 224\). The temporal and spatial patch sizes are both 28. The order of dimensions is transposed to form multi-view input volumes: axial view (\(d \times h \times w\)), sagittal view (\(h \times d \times w\)), and coronal view (\(h \times w \times d\)). Here, \(d\) stands for depth, \(h\) for height, and \(w\) for width. The top-\(k\) of~\cmke~is 16. The learning rates of the 3D visual extractor in the axial, coronal, and sagittal views, along with the other parameters, are set to $5\times10^{-5}$, $5\times10^{-5}$, $5\times10^{-5}$, and $1\times10^{-4}$, respectively. We decay the learning rate by a factor of 0.8 per epoch and set the beam size to 3. The model is trained with a batch size of 2 for 30 epochs using one NVIDIA RTX 4090 D, with Adam~\cite{kingma2014adam} as the optimizer and automatic mixed precision in PyTorch. The maximum length of generated reports is specified as 150. The words with a frequency of greater than three are retained, resulting in a vocabulary of 931 words.

\subsection{Quantitative Analysis}
To showcase the validity of the proposed \mvketr, we compare it with the recent state-of-the-art methods, including CNN-RNN(i.e., MRMA~\cite{xue2018mrma}), CNN-Transformer(i.e., Vanilla Transformer~\cite{Vaswani2017atten}, M2TR~\cite{nooralahzadeh2021m2tr}, R2GenCMN~\cite{chen-etal-2021-cross-modal}, TSGET~\cite{yi2024tsget}, UDT~\cite{yi2024udt}, LHR-RFL~\cite{yi2024lhr}, CAMANet~\cite{wang2024camanet}, SL-DG~\cite{tang2024work}), Vit-Transformer(i.e., CT2Rep~\cite{hamamci2024ct2rep}), knowledge base(i.e., M2KT~\cite{yang2023m2kt})
and large language model(i.e., Dia-LLaMA~\cite{chen2024dia}, Reg2RG~\cite{chen2025large}). As Table~\ref{tab:main_results} shows, on the CTRG-Chest-548K dataset, \mvketr~outperforms the recent SOTA approaches in almost all evaluation metrics by a considerable margin. 
Compared to MRMA, our method achieves an average gain of 17.13 across all metrics, with the most considerable improvement of 20.8 in ROUGE-L. Among CNN-Transformer methods, the vanilla Transformer yields relatively poor performance, showing a significant gap compared with our \mvketr. More recent works like M2TR and R2GenCMN show improved performance but still fall behind our method. The latest transformer-based approaches, including TSGET, UDT, LHR-RFL, and CT2Rep, achieve competitive performance by proposing specialized architectural designs or reinforcement learning strategies, while our method surpasses them all. It is worth noting that in the BLEU-1 metric, LHR-RFL achieves a second-best result.
Nevertheless, our method showcases superior performance across all evaluation metrics, with an average improvement of 4.40. M2KT achieves moderate performance, suggesting that merely incorporating a knowledge base and additional disease labels may be insufficient. By adapting the pretrained LLaMA2-7B for CT report generation, Dia-LLaMA shows promising results. However, there remains potential for further improvement. Another large language model (LLM)-based method, Reg2RG, achieves the best METEOR score of 49.71 by leveraging a region-guided referring and grounding mechanism, while our approach demonstrates superior performance in other metrics. Notably, compared to the second-best method, CAMANet, our approach achieves an average improvement of 2.51 in terms of all metrics, demonstrating the superiority of our multi-view knowledge-enhanced architecture.

\begin{table*}[!t]  
\caption{The performance comparison of our proposed method with recent SOTA works on the test set of CTRG-Chest-548K. $\ddagger$ denotes results directly cited from the original paper. $*$ represents replicated results by their public codes. The best and second-best results are highlighted in \textbf{bold} and \underline{underscored}, respectively.}
\setlength{\tabcolsep}{4mm}
\centering  
\begin{tabular}{|l|c|c|c|c|c|c|c|}  
    \hline
    Method & Year & BLEU-1 & BLEU-2 & BLEU-3 & BLEU-4 & METEOR & ROUGE-L \\ 
    \hline
    Vanilla Transformer~\cite{Vaswani2017atten}$^*$
    & 2017 &32.37 &28.4&25.82&23.93 & 22.22 & 51.28 \\
    MRMA~\cite{xue2018mrma}$^*$
    & 2018 &40.85 &30.16&23.77&19.54 & 19.52 & 33.45 \\
    M2TR~\cite{nooralahzadeh2021m2tr}$^*$
    & 2021 &39.46 &32.76&28.56&25.72 & 22.35 &47.63 \\
    R2GenCMN~\cite{chen-etal-2021-cross-modal}$^*$
    & 2022 &48.29 &38.34&32.53&28.42 & 23.89 & 48.91 \\ 
    M2KT~\cite{yang2023m2kt}$^*$
    & 2023 &39.71 &33.17&29.12&26.16 &20.11 &47.19 \\
    TSGET~\cite{yi2024tsget}$^*$
    & 2024 &46.12 &38.15&32.92&28.84 & 23.02 &50.19 \\
    UDT~\cite{yi2024udt}$^*$
    & 2024 &46.54 &39.17&34.48&31.08 & 23.42 & 53.36 \\
    CAMANet~\cite{wang2024camanet}$^*$
    & 2024 & 54.28 &\underline{45.26} &\underline{39.42} &\underline{35.38} &26.48 &\underline{54.2}  \\
    CT2Rep~\cite{hamamci2024ct2rep}$^*$
    & 2024 &49.81 &40.7&35.19&31.28 &24.82 &51.18 \\  
    SL-DG~\cite{tang2024work}$^{\ddagger}$
    & 2024 &-&--&--&23.70 & 21.90 & 43.80  \\   
    Dia-LLaMA~\cite{chen2024dia}$^{\ddagger}$
    & 2024 &51.16&--&--&29.64 & 26.28 & 42.15  \\ 
    LHR-RFL~\cite{yi2024lhr}$^*$
    & 2024 &\underline{55.56}&44.76&37.62&32.64 & 25.4 & 47.7  \\     
    Reg2RG~\cite{chen2025large}$^{\ddagger}$
    & 2024 &49.63&41.43&35.91&32.04& \bf{49.71} & 47.76  \\ 
\mvketr(Ours)
    & -- & \bf{58.36}& \bf{48.79}  &\bf{42.43}  &\bf{37.86} &\underline{28.36} &\bf{54.25}\\
    \hline
        \multicolumn{8}{p{200pt}}{All evaluation metrics are illustrated as percentage (\%).}\\
\end{tabular}  
\label{tab:main_results}
\end{table*}

\begin{table*}[!t]
\caption{Results of ablation experiments. The average performance improvement over all metrics compared to BASE is presented in the ``AVG.$\Delta$" column. The best results are highlighted in \textbf{bold}.}
\setlength{\tabcolsep}{4mm}
\centering
\begin{tabular}{|l|c|c|c|c|c|c|c|}
    \hline
    Method & BLEU-1 & BLEU-2 & BLEU-3 & BLEU-4 & METEOR & ROUGE-L & AVG.$\Delta$\\
    \hline
    BASE      & 48.29 &38.34&32.53&28.42 & 23.89 & 48.91 & -\\ 
    BASE+\mvpa~& 54.81 & 45.82 & 39.85 & 35.49 & 26.54 & \textbf{54.68} & 17.2\% \\ 
    BASE+\cmke~& 53.76 & 44.61 & 38.63 & 34.43 & 26.77 & 52.59 &14.5\%\\ 
   Ours-MLP  & 53.98& 45.3 &39.55  &35.39 &27.16&54.05&14.1\%\\
    Ours  & \bf{58.36}& \bf{48.79}  &\bf{42.43}  &\bf{37.86} &\bf{28.36} &54.25&\bf{23.6\%}\\
    \hline
        \multicolumn{8}{p{200pt}}{All evaluation metrics are illustrated as percentage (\%).}\\
\end{tabular}
\label{table:ablation}
\end{table*}

\subsection{Qualitative Analysis}
\begin{figure*}[!t]
\centering
\includegraphics[width=\textwidth]{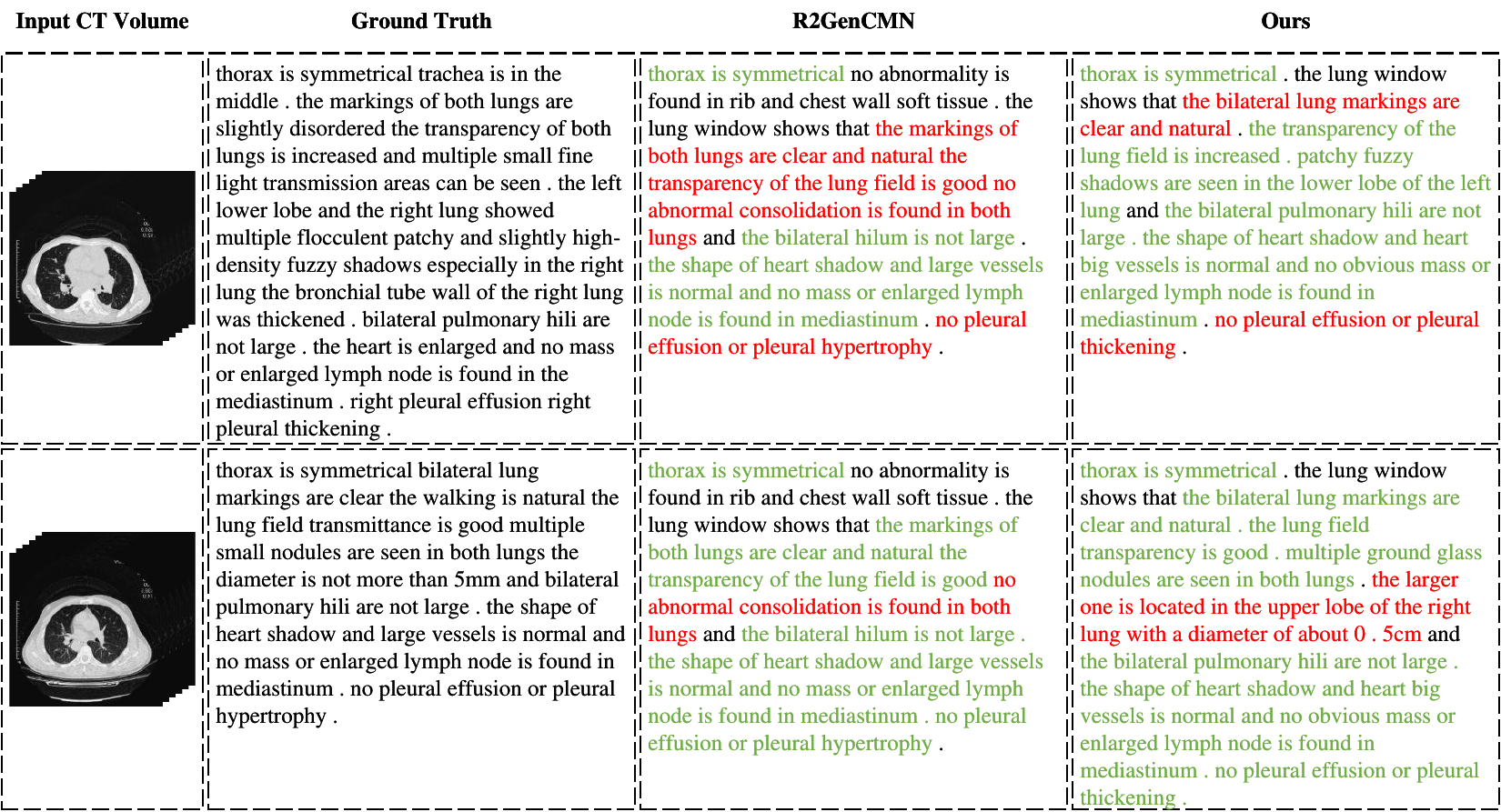} 
\caption{The examples of report from ground-truth, R2GenCMN, and our method. \textcolor{mygreen}{Green}/\textcolor{red}{red} highlights denotes correct/incorrect content respectively.}  
\label{fig:result}    
\end{figure*}

In addition to quantitative analysis, we also conduct qualitative analysis on test cases by comparing their ground truths with the reports generated from our proposed method, \mvketr, and its baseline, R2GenCMN, upon which our approach is built. Fig.~\ref{fig:result} shows two examples from CTRG-Chest-548K and their corresponding reports, where green and red highlights denote correct and incorrect content, respectively. In both cases, \mvketr~generates reports that resemble radiologists' observations in terms of clinical accuracy and professional terminology. To better illustrate the performance difference between \mvketr~and R2GenCMN, a detailed analysis of these two cases is presented below. 

In the first case(top row of Fig.\ref{fig:result}), \mvketr~accurately detects key abnormalities such as increased lung transparency and patchy shadows in the left lower lobe. Although both methods make some errors in describing pleural effusion/thickening, \mvketr~provides a more comprehensive and accurate description similar to the ground truth report. For the second case (bottom row of Fig.\ref{fig:result}), \mvketr~successfully spots the presence of multiple nodules in both lungs, which is a critical diagnostic finding that R2GenCMN completely overlooks. These two cases suggest that \mvketr~not only provides more accurate and comprehensive reports but also shows better capability in detecting clinically significant abnormalities, which is essential for reliable medical diagnosis.

\begin{table*}[!t]
\caption{human evaluation by two board-certified radiologists (5-point scale: 1=Reject; 5=Accept). Scores were averaged across 10 clinical cases.  The best and second-best results are highlighted in \textbf{bold} and \underline{underscored}, respectively.}
\setlength{\tabcolsep}{4mm}
\centering
\begin{tabular}{|l|c|c|c|c|}
    \hline
    Method & completeness & diagnostic precision & readability & diagnostic confidence \\
    \hline
     MRMA~\cite{xue2018mrma}& 3.45 & 1.7 & 1.75 & 1.75 \\ 
    M2KT~\cite{yang2023m2kt} & \underline{3.7} & \underline{2.8} & \underline{3.05} & \underline{2.95}\\ 
    M2TR~\cite{nooralahzadeh2021m2tr} & 2.8 & 2.05 & 2.05 & 2 \\ 
    Ours & \bf{4.2}& \bf{3.9}  &\bf{4.15}  &\bf{4.1}\\
    \hline
\end{tabular}
\label{table:human_evaluation}
\end{table*}

\subsection{Human Evaluation}
To further validate the clinical utility of our method, we conduct human evaluations. Two board-certified radiologists from a leading tertiary care hospital in China (8–15 years of experience) were invited to independently assess the reports from four clinical quality dimensions: completeness (coverage of all key clinical findings, including lesion location, size, density characteristics, and absence of critical information omissions), diagnostic precision (ability to differentiate ambiguous lesions), readability (logical flow, terminology accuracy, and clarity), and diagnostic confidence (likelihood of adopting the generated report in clinical practice). They are required to give an overall 5-scale grade (5: Accept, 4: Weakly Accept, 3: Borderline, 2: Weakly Reject, 1: Reject). This hospital ranks among the nation's top 50 hospitals and processes approximately 250,000 to 300,000 chest CT examinations annually, ensuring clinically representative expertise in thoracic imaging. We randomly select 10 chest CT scans from the testing set of CTRG-Chest-548K and provide their corresponding reports generated by MRMA~\cite{xue2018mrma}, M2KT~\cite{yang2023m2kt}, M2TR~\cite{nooralahzadeh2021m2tr}, and our model for the radiologists. They are unaware of which model generates these reports and are encouraged to select a higher-quality report among the four methods based on the predefined clinical dimensions. As depicted in Table~\ref{table:human_evaluation}, our method surpasses all comparative approaches with significant margins across four clinical quality metrics. Notably, our approach achieves the highest scores in \textit{completeness} (4.2), \textit{diagnostic precision} (3.9), \textit{readability} (4.15), and \textit{diagnostic confidence} (4.1), yielding an average improvement of 32\% over the second-best method, M2KT. This significant margin highlights that our model generates clinically reliable and radiologist-preferred reports, directly addressing concerns about practical relevance and diagnostic accuracy in radiology workflows.

\subsection{Ablation Study}

\begin{table*}[!t]
\caption{The performance comparison of the \mvketr~framework under different 3D visual extractor. The best results are highlighted in \textbf{bold}.}
\setlength{\tabcolsep}{4mm}
\centering
\begin{tabular}{|l|c|c|c|c|c|c|}
    \hline
    Visual Extractor & BLEU-1 & BLEU-2 & BLEU-3 & BLEU-4 & METEOR & ROUGE-L\\
    \hline
     3D ViT~\cite{dosovitskiy2021vit}& \bf{58.39} & 47.27 & 39.87 & 34.59 & 26.99 & 49.43\\ 
    CT-Net~\cite{draelos2021ctnet} & 50.1 & 42.24 & 37.03 & 33.33&26.17 &  53.67\\ 
    U-Net~\cite{ronneberger2015unet} & 48.55 & 39.87 & 34.72 &31.1  &  24.75&52.27 \\ 
    CT-ViT~\cite{hamamci2024generatect} & 58.36& \bf{48.79}  &\bf{42.43}  &\bf{37.86} &\bf{28.36} &\bf{54.25}\\
    \hline
        \multicolumn{7}{p{200pt}}{All evaluation metrics are illustrated as percentage (\%).}\\
\end{tabular}
\label{table:ve_compare}
\end{table*}

\begin{table*}[!t]
\caption{Performance analysis of view-aware attention under varying rotation angles and suboptimal imaging conditions. The average performance change across all metrics compared to the baseline (our proposed \mvketr) is presented in the ``AVG.$\Delta$" column, with positive values indicating improvement and negative values indicating degradation. The best results in each scenario group are highlighted in \textbf{bold}.}
\setlength{\tabcolsep}{3mm}
\centering
    \begin{tabular}{|l|l|c|c|c|c|c|c|c|}
    \hline
        Scenario & Type & BLEU-1 & BLEU-2 & BLEU-3 & BLEU-4 & METEOR & ROUGE-L & AVG.$\Delta$ \\ \hline
        Baseline & Original & 58.36 & 48.79 & 42.43 & 37.86 & 28.36 & 54.25 & - \\ \hline
        \multirow{4}{*}{\begin{tabular}[c]{@{}l@{}}Rotation\\Angle\end{tabular}} & ±5° & 55.91  & 46.62  & 40.46  & 36.00  & 27.20  & 54.06  & -3.80\%  \\
        & ±10° & \bf{56.60}  & 47.27  & 41.12  & 36.69  & 27.57  & 54.34  & -2.50\%  \\
        & ±15° & 55.69  & 46.94  & 41.12  & \bf{36.91}  & \bf{27.80}  & 54.42  & -2.60\%  \\
        & ±20° & 56.44  & \bf{47.30}  & \bf{41.21}  & 36.84  & 27.62  & \bf{54.70}  & \bf{-2.30\%} \\ \hline
        \multirow{3}{*}{\begin{tabular}[c]{@{}l@{}}Imaging\\Condition\end{tabular}} & Motion Artifacts~\cite{buzug2011computed} & 58.15  & 48.94  & 42.82  & \bf{38.40}  & 28.51  & \bf{55.21}  & \bf{+0.80\%}  \\
        & Ring Artifacts~\cite{buzug2011computed} & \bf{59.78}  & \bf{49.58}  & \bf{42.90}  & 38.13  & \bf{28.53}  & 53.03  & +0.70\%  \\
        & MPR Blurring~\cite{remy1998multiplanar} & 57.79  & 48.53  & 42.37  & 37.95  & 28.33  & 53.94 & -0.30\%  \\ \hline
        \multicolumn{9}{p{200pt}}{All evaluation metrics are illustrated as percentage (\%).}\\        
    \end{tabular}
\label{table:limitations_vva}    
\end{table*}

Furthermore, we perform a body of ablation experiments to investigate the contributions of each component. The following baselines are used:

\begin{itemize}
\item BASE: The base model employs CT-ViT as the 3D visual extractor and the encoder-decoder of R2GenCMN as the report generator.
\item BASE+\mvpa: The base model enhanced with the proposed multi-view perception aggregator.
\item BASE+\cmke: The base model integrated with the proposed cross-modal knowledge enhancer.
\item Ours-MLP: This is the complete model with KAN layers replaced by MLP layers.
\item Ours: This is the complete model with all proposed components.
\end{itemize}

As shown in Table~\ref{table:ablation}, we can draw several key observations. First, compared to BASE, BASE+\mvpa~achieves an average improvement of 17.2\%, which confirms the efficacy of integrating diagnostic information across multiple anatomical views through view-aware attention. Second, BASE+\cmke~shows a 14.5\% average improvement over BASE, indicating the usefulness of incorporating medical expertise from similar cases. The 14.5\% BLEU-4 degradation without CMKE confirms that retrieved knowledge primarily enhances model generalization instead of replacing intrinsic visual reasoning capabilities. Particularly, BASE+\mvpa~achieves better performance than BASE+\cmke. This observation is attributed to the fact that multi-view perception is more crucial than clinical knowledge enhancement. Third, the comparison between Ours-MLP and Ours reveals a performance gap of 9.5\% (23.6\% vs 14.1\% in average improvement). This observation owes to the fact that KAN layers are more effective in modeling complicated diagnostic relationships compared to conventional MLP layers.

When combining all components, Ours achieves the best overall performance, with a 23.6\% average improvement over BASE. This suggests that multi-view perception, knowledge enhancement, and the KAN network work complementarily to improve the quality of CT reports.

\begin{figure}[!t]
\centering
\includegraphics[width=0.4\textwidth]{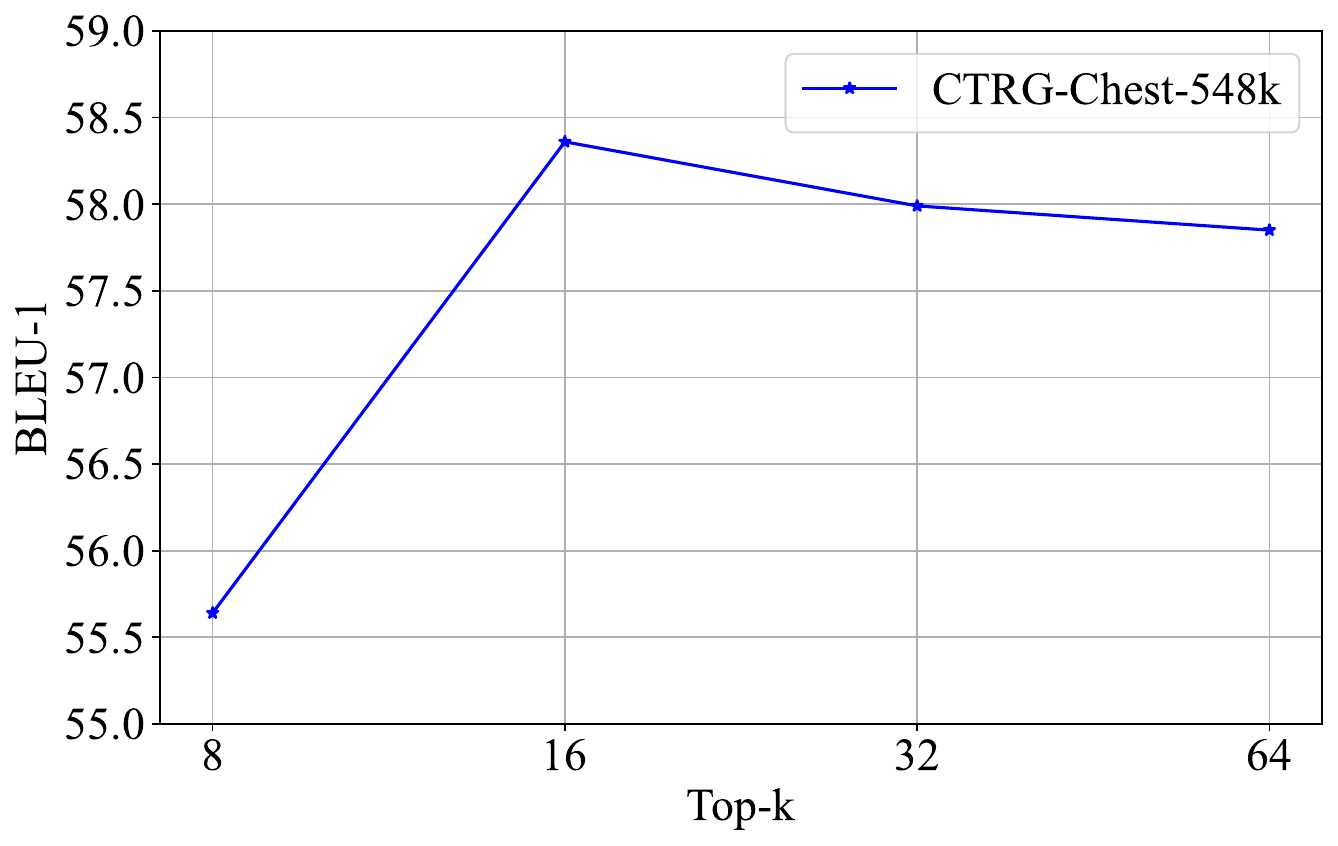}  
\caption{Effect of varying top-k on CTRG-Chest-548K.}  
\label{fig:effect_top_k}    
\end{figure}

\subsection{Discussion}
\subsubsection{Comparison of different 3D visual extractors}
We conduct experiments to assess the impact of various 3D visual extractors, including CNN-based methods: U-Net~\cite{ronneberger2015unet}, and CT-Net~\cite{draelos2021ctnet} as well as recent Vit-based methods: 3D ViT~\cite{dosovitskiy2021vit}, and CT-ViT~\cite{hamamci2024generatect}. As demonstrated in Table~\ref{table:ve_compare}, we find that CT-ViT(see Fig.~\ref{fig:pipeline}) is the most suitable 3D visual extractor for our proposed \mvketr, which leverages the advantages of two types of transformers: spatial transformers and causal transformers. Specifically, the spatial transformers excel at capturing local spatial features and relationships within CT slices, which is crucial for understanding anatomical structures and lesion locations. Meanwhile, the causal transformers are adept at modeling long-range dependencies across different slices or depth levels of the 3D CT volume, ensuring comprehensive feature extraction. This combination effectively preserves spatial and volumetric information throughout the 3D visual feature extraction pipeline. Moreover, the ViT-based methods show relatively higher performance compared to the CNN-based counterparts, suggesting that the self-attention mechanism is more effective in capturing global dependencies and long-range interactions within visual features, which is crucial for CT report generation.

\subsubsection{Effect of varying top-k}

The top-$k$ is a key hyperparameter of \cmke~module, which determines the number of most relevant reports to be retrieved for the current report generation. As revealed in Fig.~\ref{fig:effect_top_k}, we observe that increasing the top-$k$ at first improves the BLEU-1 score, reaching a peak at top-$k$=16, beyond which the performance deteriorates. This trend suggests that while retrieving more relevant reports provides richer reference information, excessive retrieval beyond a certain threshold may introduce noise without additional benefits. Hence, the top-$k$ is set to 16, balancing noise suppression and information retention.

\subsubsection{Limitations and Future Work}
Notwithstanding our contributions to automatic radiology report generation, there are still limitations. We have conducted comprehensive experiments to evaluate the robustness of our proposed MvKeTR (the baseline model, as clearly labeled in Table~\ref{table:limitations_vva}) under challenging scenarios. First, to analyze the performance of view-aware attention under varying rotation angles, we conducted controlled experiments on the CTRG-Chest-548K dataset using rotations (±5°, ±10°, ±15°, and ±20°). As shown in Table~\ref{table:limitations_vva}, we observed an interesting nonlinear degradation pattern. While the smallest rotation (±5°) resulted in the largest average performance drop (3.80\%), larger rotations exhibited more moderate degradation (2.50\%, 2.60\%, and 2.30\% for ±10°, ±15°, and ±20°, respectively). This counterintuitive finding suggests that MVPA dynamically adapts to misalignment severity, possibly by redistributing attention weights across views. The detailed metrics reveal that certain performance indicators (e.g., METEOR at ±15° and ROUGE-L at ±20°) remain robust even under significant misalignment, demonstrating the architecture's ability to leverage multi-view redundancy. This adaptive behavior indicates that MVPA suppresses features from severely misaligned views while amplifying more stable features from other perspectives, resulting in a graceful degradation pattern rather than catastrophic failure. Controlled rotational perturbations during training empower MVPA to learn alignment-invariant representations while preserving its adaptive attention mechanism, thereby ensuring robust generalization to real-world scenarios with inevitable misalignment.

Second, we further evaluated our method under three common suboptimal imaging conditions where anatomical view discrimination becomes challenging: motion artifacts causing slice discontinuity~\cite{buzug2011computed}, ring artifacts from CT detector miscalibration~\cite{buzug2011computed}, and multiplanar reconstruction (MPR) blurring at oblique angles~\cite{remy1998multiplanar}. Surprisingly, as shown in Table~\ref{table:limitations_vva}, our model maintains performance and demonstrates slight improvement under certain conditions (e.g., motion artifacts). For motion artifacts that induce slice discontinuity and view misregistration, the model achieved a 0.80\% average performance improvement, as the axial view (less affected by respiratory motion) provided stable features that compensated for corrupted coronal/sagittal views. In cases of ring artifacts, the average performance improved by 0.70\%. Since ring artifacts predominantly affect axial slices, the view-aware attention mechanism naturally shifted focus to coronal/sagittal planes where anatomical continuity remained intact, replicating radiologists' cross-plane verification strategies. Under MPR blurring, the negligible 0.30\% performance drop reflects MVPA's tolerance to resolution variations, achieved through its multi-scale feature fusion design that combines native-plane high-frequency components with cross-view contextual representations.
These findings collectively validate that view-aware attention functions as an adaptive spatial filter rather than a rigid view selector. By dynamically adjusting cross-view feature contributions based on anatomical coherence and imaging quality, MVPA emulates clinicians' diagnostic reasoning under uncertainty. The results confirm our architecture's clinical viability in real-world scenarios where perfect alignment and optimal imaging conditions are often unattainable.
These limitations underscore two strategic extensions for future research: 1) Integration of CT physics-informed artifact models (e.g., low-dose Poisson noise propagation) to enhance domain shift robustness; 2) Geometric adaptation of the view-aware attention mechanism for non-orthogonal multiplanar reconstructions, directly addressing emerging clinical requirements in volumetric imaging analysis.

In addition to the aforementioned limitations in the view-aware attention mechanism, our CMKE also presents challenges. While CMKE demonstrates robustness to noisy reports, its performance relies on the quality of the CT-RATE dataset. To further enhance reliability, we plan to introduce:  
1) Uncertainty-aware retrieval: Filtering reports with low similarity scores (e.g., below a threshold) to exclude low-confidence cases.  
2) Human-in-the-loop validation: Enabling radiologists to review and adjust retrieved reports during inference.  
3) Bias mitigation: Training CT-CLIP with debiasing techniques (e.g., adversarial training) to reduce dataset-specific biases.  
These improvements will strengthen the alignment between retrieved knowledge and clinical ground truth.

To bridge the gap between technical advancements and clinical deployment, we will focus on validating the generalizability and effectiveness of our proposed model through a real-world pilot study across local hospitals and extending the framework to other 3D medical imaging modalities (e.g., MRI) and different body parts.

\section{Conclusion}
\label{sec:conclusion}
In this paper, we present \mvketr, a novel framework that integrates multi-view perception and knowledge enhancement for high-quality radiology report generation from 3D CT volumes. First, multi-view features are extracted by a 3D visual extractor, which comprises three separate CT-ViT networks. The multi-view perception aggregator is then utilized to synthesize visual features from three anatomical views. Following this, the cross-modal knowledge enhancer is responsible for incorporating clinical knowledge from the most relevant cases into the diagnostic process. Finally, these features encoded by the two aforementioned modules are concatenated and fed into the report generator to produce the final report. Experimental results on the CTRG-Chest-548K dataset demonstrate the superiority of our method over prior state-of-the-art studies.
\section*{References}
\bibliography{IEEEabrv,refs}

\begin{thebibliography}{10}
\providecommand{\url}[1]{#1}
\csname url@samestyle\endcsname
\providecommand{\newblock}{\relax}
\providecommand{\bibinfo}[2]{#2}
\providecommand{\BIBentrySTDinterwordspacing}{\spaceskip=0pt\relax}
\providecommand{\BIBentryALTinterwordstretchfactor}{4}
\providecommand{\BIBentryALTinterwordspacing}{\spaceskip=\fontdimen2\font plus
\BIBentryALTinterwordstretchfactor\fontdimen3\font minus \fontdimen4\font\relax}
\providecommand{\BIBforeignlanguage}[2]{{%
\expandafter\ifx\csname l@#1\endcsname\relax
\typeout{** WARNING: IEEEtran.bst: No hyphenation pattern has been}%
\typeout{** loaded for the language `#1'. Using the pattern for}%
\typeout{** the default language instead.}%
\else
\language=\csname l@#1\endcsname
\fi
#2}}
\providecommand{\BIBdecl}{\relax}
\BIBdecl

\bibitem{goergen2013evidence}
S.~K. Goergen, F.~J. Pool, T.~J. Turner, J.~E. Grimm, M.~N. Appleyard, C.~Crock, M.~C. Fahey, M.~F. Fay, N.~J. Ferris, S.~M. Liew \emph{et~al.}, ``Evidence-based guideline for the written radiology report: Methods, recommendations and implementation challenges,'' \emph{J. Med. Imaging Radiat. Oncol.}, vol.~57, no.~1, pp. 1--7, 2013.

\bibitem{rosenkrantz2016us}
A.~B. Rosenkrantz, D.~R. Hughes, and R.~Duszak~Jr, ``The us radiologist workforce: an analysis of temporal and geographic variation by using large national datasets,'' \emph{Radiology}, vol. 279, no.~1, pp. 175--184, 2016.

\bibitem{rimmer2017radiologist}
A.~Rimmer, ``Radiologist shortage leaves patient care at risk, warns royal college,'' \emph{BMJ-BRIT. MED. J.}, vol. 359, 2017.

\bibitem{bruno2015understanding}
M.~A. Bruno, E.~A. Walker, and H.~H. Abujudeh, ``Understanding and confronting our mistakes: the epidemiology of error in radiology and strategies for error reduction,'' \emph{Radiographics}, vol.~35, no.~6, pp. 1668--1676, 2015.

\bibitem{chen-etal-2021-cross-modal}
Z.~Chen, Y.~Shen, Y.~Song, and X.~Wan, ``Cross-modal memory networks for radiology report generation,'' in \emph{Proc. 59th Ann. Meeting Assoc. Comput. Linguistics, 11th Int. Joint Conf. Natural Lang. Process}, 2021, pp. 5904--5914.

\bibitem{jin2024improving}
Y.~Jin, W.~Chen, Y.~Tian, Y.~Song, C.~Yan, and Z.~Mao, ``Improving radiology report generation with d 2-net: When diffusion meets discriminator,'' in \emph{Proc. IEEE Int. Conf. Acoust. Speech Signal Process.}, 2024, pp. 2215--2219.

\bibitem{muller2002computed}
N.~M{\"u}ller, ``Computed tomography and magnetic resonance imaging: past, present and future,'' \emph{Eur. Respir. J.}, vol.~19, no. 35 suppl, pp. 3s--12s, 2002.

\bibitem{tang2024work}
Y.~Tang, H.~Yang, L.~Zhang, and Y.~Yuan, ``Work like a doctor: Unifying scan localizer and dynamic generator for automated computed tomography report generation,'' \emph{Expert Syst. Appl.}, vol. 237, p. 121442, 2024.

\bibitem{hamamci2024ct2rep}
I.~E. Hamamci, S.~Er, and B.~Menze, ``Ct2rep: Automated radiology report generation for 3d medical imaging,'' in \emph{Proc. Int. Conf. Med. Image Comput. Comput.-Assisted Intervention}, 2024, pp. 476--486.

\bibitem{chen2024dia}
Z.~Chen, L.~Luo, Y.~Bie, and H.~Chen, ``Dia-llama: Towards large language model-driven ct report generation,'' \emph{arXiv:2403.16386}, 2024.

\bibitem{setio2016pulmonary}
A.~A.~A. Setio, F.~Ciompi, G.~Litjens, P.~Gerke, C.~Jacobs, S.~J. Van~Riel, M.~M.~W. Wille, M.~Naqibullah, C.~I. S{\'a}nchez, and B.~Van~Ginneken, ``Pulmonary nodule detection in ct images: false positive reduction using multi-view convolutional networks,'' \emph{{IEEE} Trans. Med. Imaging}, vol.~35, no.~5, pp. 1160--1169, 2016.

\bibitem{bankier2017recommendations}
A.~A. Bankier, H.~MacMahon, J.~M. Goo, G.~D. Rubin, C.~M. Schaefer-Prokop, and D.~P. Naidich, ``Recommendations for measuring pulmonary nodules at ct: a statement from the fleischner society,'' \emph{Radiology}, vol. 285, no.~2, pp. 584--600, 2017.

\bibitem{macmahon2017guidelines}
H.~MacMahon, D.~P. Naidich, J.~M. Goo, K.~S. Lee, A.~N. Leung, J.~R. Mayo, A.~C. Mehta, Y.~Ohno, C.~A. Powell, M.~Prokop \emph{et~al.}, ``Guidelines for management of incidental pulmonary nodules detected on ct images: from the fleischner society 2017,'' \emph{Radiology}, vol. 284, no.~1, pp. 228--243, 2017.

\bibitem{chen-emnlp-2020-r2gen}
Z.~Chen, Y.~Song, T.-H. Chang, and X.~Wan, ``Generating radiology reports via memory-driven transformer,'' in \emph{Proc. Conf. Empirical Methods Natural Lang. Process.}, Nov. 2020, pp. 1439--1449.

\bibitem{liu2025kan}
Z.~Liu, Y.~Wang, S.~Vaidya, F.~Ruehle, J.~Halverson, M.~Soljacic, T.~Y. Hou, and M.~Tegmark, ``{KAN}: Kolmogorov{\textendash}arnold networks,'' in \emph{Proc. 13th Int. Conf. Learn. Represent.}, 2025.

\bibitem{bai2018survey}
S.~Bai and S.~An, ``A survey on automatic image caption generation,'' \emph{Neurocomputing}, vol. 311, pp. 291--304, 2018.

\bibitem{wang2020overview}
H.~Wang, Y.~Zhang, and X.~Yu, ``An overview of image caption generation methods,'' \emph{Comput. Intell. Neurosci.}, vol. 2020, no.~1, p. 3062706, 2020.

\bibitem{yao2010i2t}
B.~Z. Yao, X.~Yang, L.~Lin, M.~W. Lee, and S.-C. Zhu, ``I2t: Image parsing to text description,'' \emph{Proc. IEEE}, vol.~98, no.~8, pp. 1485--1508, 2010.

\bibitem{socher2010connecting}
R.~Socher and L.~Fei-Fei, ``Connecting modalities: Semi-supervised segmentation and annotation of images using unaligned text corpora,'' in \emph{Proc. IEEE Conf. Comput. Vis. Pattern Recognit.}, 2010, pp. 966--973.

\bibitem{vinyals2015show}
O.~Vinyals, A.~Toshev, S.~Bengio, and D.~Erhan, ``Show and tell: A neural image caption generator,'' in \emph{Proc. IEEE Conf. Comput. Vis. Pattern Recognit.}, 2015, pp. 3156--3164.

\bibitem{rennie2017self}
S.~J. Rennie, E.~Marcheret, Y.~Mroueh, J.~Ross, and V.~Goel, ``Self-critical sequence training for image captioning,'' in \emph{Proc. IEEE Conf. Comput. Vis. Pattern Recognit.}, 2017, pp. 7008--7024.

\bibitem{xu2015show}
K.~Xu, ``Show, attend and tell: Neural image caption generation with visual attention,'' \emph{arXiv:1502.03044}, 2015.

\bibitem{li2017image}
L.~Li, S.~Tang, L.~Deng, Y.~Zhang, and Q.~Tian, ``Image caption with global-local attention,'' in \emph{Proc. AAAI Conf. Artif. Intell.}, vol.~31, no.~1, 2017.

\bibitem{lu2017knowing}
J.~Lu, C.~Xiong, D.~Parikh, and R.~Socher, ``Knowing when to look: Adaptive attention via a visual sentinel for image captioning,'' in \emph{Proc. IEEE Conf. Comput. Vis. Pattern Recognit.}, 2017, pp. 375--383.

\bibitem{cornia2020meshed}
M.~Cornia, M.~Stefanini, L.~Baraldi, and R.~Cucchiara, ``Meshed-memory transformer for image captioning,'' in \emph{Proc. IEEE Conf. Comput. Vis. Pattern Recognit.}, 2020, pp. 10\,578--10\,587.

\bibitem{liu2021cptr}
W.~Liu, S.~Chen, L.~Guo, X.~Zhu, and J.~Liu, ``Cptr: Full transformer network for image captioning,'' \emph{arXiv:2101.10804}, 2021.

\bibitem{jing2017automatic}
B.~Jing, P.~Xie, and E.~Xing, ``On the automatic generation of medical imaging reports,'' \emph{arXiv:1711.08195}, 2017.

\bibitem{xue2018mrma}
Y.~Xue, T.~Xu, L.~Rodney~Long, Z.~Xue, S.~Antani, G.~R. Thoma, and X.~Huang, ``Multimodal recurrent model with attention for automated radiology report generation,'' in \emph{Proc. 21st Int. Conf. Med. Image Comput. Comput.-Assisted Intervention}, 2018, pp. 457--466.

\bibitem{hoogi2020natural}
A.~Hoogi, A.~Mishra, F.~Gimenez, J.~Dong, and D.~Rubin, ``Natural language generation model for mammography reports simulation,'' \emph{{IEEE} J. Biomed. Health Inform.}, vol.~24, no.~9, pp. 2711--2717, 2020.

\bibitem{wang2021self}
Z.~Wang, L.~Zhou, L.~Wang, and X.~Li, ``A self-boosting framework for automated radiographic report generation,'' in \emph{Proc. IEEE Conf. Comput. Vis. Pattern Recognit.}, 2021, pp. 2433--2442.

\bibitem{nooralahzadeh2021m2tr}
F.~Nooralahzadeh, N.~Perez~Gonzalez, T.~Frauenfelder, K.~Fujimoto, and M.~Krauthammer, ``Progressive transformer-based generation of radiology reports,'' in \emph{Proc. Findings Assoc. Comput. Linguistics: EMNLP 2021}, 2021, pp. 2824--2832.

\bibitem{yi2024tsget}
X.~Yi, Y.~Fu, R.~Liu, H.~Zhang, and R.~Hua, ``Tsget: Two-stage global enhanced transformer for automatic radiology report generation,'' \emph{{IEEE} J. Biomed. Health Inform.}, vol.~28, no.~4, pp. 2152--2162, 2024.

\bibitem{yi2024udt}
X.~Yi, Y.~Fu, R.~Hua, R.~Liu, and H.~Zhang, ``Unsupervised disease tags for automatic radiology report generation,'' \emph{Biomed Signal Process Control}, vol.~89, p. 105742, 2024.

\bibitem{yi2024lhr}
X.~Yi, Y.~Fu, J.~Yu, R.~Liu, H.~Zhang, and R.~Hua, ``Lhr-rfl: Linear hybrid-reward based reinforced focal learning for automatic radiology report generation,'' \emph{{IEEE} Trans. Med. Imaging}, 2024.

\bibitem{wang2024camanet}
J.~Wang, A.~Bhalerao, T.~Yin, S.~See, and Y.~He, ``Camanet: Class activation map guided attention network for radiology report generation,'' \emph{{IEEE} J. Biomed. Health Inform.}, vol.~28, no.~4, pp. 2199--2210, 2024.

\bibitem{zhang2023weakly}
X.~Zhang, S.~Yang, Y.~Shi, J.~Ji, Y.~Liu, Z.~Wang, and H.~Xu, ``Weakly guided attention model with hierarchical interaction for brain ct report generation,'' \emph{Comput. Biol. Med.}, vol. 167, p. 107650, 2023.

\bibitem{zhang2024co}
X.~Zhang, S.~Dou, J.~Ji, Y.~Liu, and Z.~Wang, ``Co-occurrence relationship driven hierarchical attention network for brain ct report generation,'' \emph{{IEEE} Trans. Emerging Top. Comput. Intell.}, 2024.

\bibitem{touvron2023llama}
H.~Touvron, L.~Martin, K.~Stone, P.~Albert, A.~Almahairi, Y.~Babaei, N.~Bashlykov, S.~Batra, P.~Bhargava, S.~Bhosale \emph{et~al.}, ``Llama 2: Open foundation and fine-tuned chat models,'' \emph{arXiv:2307.09288}, 2023.

\bibitem{chen2025large}
Z.~Chen, Y.~Bie, H.~Jin, and H.~Chen, ``Large language model with region-guided referring and grounding for ct report generation,'' \emph{{IEEE} Trans. Med. Imaging}, 2025.

\bibitem{xia2024automatic}
Z.~Xia, M.~Liao, S.~Di, Y.~Zhao, W.~Liang, and N.~N. Xiong, ``Automatic liver segmentation from ct volumes based on multi-view information fusion and condition random fields,'' \emph{Optics \& Laser Technology}, vol. 179, p. 111298, 2024.

\bibitem{liu2024multi}
C.~Liu, H.~Liu, X.~Zhang, J.~Guo, and P.~Lv, ``Multi-scale and multi-view network for lung tumor segmentation,'' \emph{Comput. Biol. Med.}, vol. 172, p. 108250, 2024.

\bibitem{yuan2019automatic}
J.~Yuan, H.~Liao, R.~Luo, and J.~Luo, ``Automatic radiology report generation based on multi-view image fusion and medical concept enrichment,'' in \emph{Proc. 22nd Int. Conf. Med. Image Comput. Comput.-Assisted Intervention}, 2019, pp. 721--729.

\bibitem{yang2020automatic}
S.~Yang, J.~Niu, J.~Wu, and X.~Liu, ``Automatic medical image report generation with multi-view and multi-modal attention mechanism,'' in \emph{International Conference on Algorithms and Architectures for Parallel Processing}.\hskip 1em plus 0.5em minus 0.4em\relax Springer, 2020, pp. 687--699.

\bibitem{yang2023m2kt}
S.~Yang, X.~Wu, S.~Ge, Z.~Zheng, S.~K. Zhou, and L.~Xiao, ``Radiology report generation with a learned knowledge base and multi-modal alignment,'' \emph{Med. Image Anal.}, vol.~86, p. 102798, 2023.

\bibitem{kale2023knowledge}
K.~Kale, P.~Bhattacharyya, A.~Shetty, M.~Gune, K.~Shrivastava, R.~Lawyer, and S.~Biswas, ``“knowledge is power”: Constructing knowledge graph of abdominal organs and using them for automatic radiology report generation,'' in \emph{Proc. 61st Ann. Meeting Assoc. Comput. Linguistics (Vol. 5: Industry Track)}, 2023, pp. 11--24.

\bibitem{hou2023mkcl}
X.~Hou, Z.~Liu, X.~Li, X.~Li, S.~Sang, and Y.~Zhang, ``Mkcl: medical knowledge with contrastive learning model for radiology report generation,'' \emph{J. Biomed. Inf.}, vol. 146, p. 104496, 2023.

\bibitem{huang2023kiut}
Z.~Huang, X.~Zhang, and S.~Zhang, ``Kiut: Knowledge-injected u-transformer for radiology report generation,'' in \emph{Proc. IEEE Conf. Comput. Vis. Pattern Recognit.}, 2023, pp. 19\,809--19\,818.

\bibitem{rumelhart1986learning}
D.~E. Rumelhart, G.~E. Hinton, and R.~J. Williams, ``Learning representations by back-propagating errors,'' \emph{nature}, vol. 323, no. 6088, pp. 533--536, 1986.

\bibitem{hornik1989multilayer}
K.~Hornik, M.~Stinchcombe, and H.~White, ``Multilayer feedforward networks are universal approximators,'' \emph{Neural networks}, vol.~2, no.~5, pp. 359--366, 1989.

\bibitem{wang2025on}
Y.~Wang, J.~W. Siegel, Z.~Liu, and T.~Y. Hou, ``On the expressiveness and spectral bias of {KAN}s,'' in \emph{Proc. 13th Int. Conf. Learn. Represent.}, 2025.

\bibitem{rahaman2019spectral}
N.~Rahaman, A.~Baratin, D.~Arpit, F.~Draxler, M.~Lin, F.~Hamprecht, Y.~Bengio, and A.~Courville, ``On the spectral bias of neural networks,'' in \emph{Proc. Int. Conf. Mach. Learn.}\hskip 1em plus 0.5em minus 0.4em\relax PMLR, 2019, pp. 5301--5310.

\bibitem{Vaswani2017atten}
A.~Vaswani, N.~Shazeer, N.~Parmar, J.~Uszkoreit, L.~Jones, A.~N. Gomez, L.~Kaiser, and I.~Polosukhin, ``Attention is all you need,'' in \emph{Proc. 31st Int. Conf. Neural Inf. Process. Syst.}, 2017, p. 6000–6010.

\bibitem{hamamci2024ct-clip}
I.~E. Hamamci, S.~Er, F.~Almas, A.~G. Simsek, S.~N. Esirgun, I.~Dogan, M.~F. Dasdelen, B.~Wittmann, E.~Simsar, M.~Simsar \emph{et~al.}, ``A foundation model utilizing chest ct volumes and radiology reports for supervised-level zero-shot detection of abnormalities,'' \emph{arXiv:2403.17834}, 2024.

\bibitem{yan2022radbert}
A.~Yan, J.~McAuley, X.~Lu, J.~Du, E.~Y. Chang, A.~Gentili, and C.-N. Hsu, ``Radbert: adapting transformer-based language models to radiology,'' \emph{Radiol. Artif. Intell.}, 2022.

\bibitem{papineni-etal-2002-bleu}
K.~Papineni, S.~Roukos, T.~Ward, and W.-J. Zhu, ``{B}leu: a method for automatic evaluation of machine translation,'' in \emph{Proc. 40th Annu. Meeting Assoc. Comput. Linguistics}, Jul. 2002, pp. 311--318.

\bibitem{denkowski2011meteor}
M.~Denkowski and A.~Lavie, ``Meteor 1.3: Automatic metric for reliable optimization and evaluation of machine translation systems,'' in \emph{Proc. 6th Workshop Stat. Mach. Transl.}, 2011, pp. 85--91.

\bibitem{lin-2004-rouge}
C.-Y. Lin, ``{ROUGE}: A package for automatic evaluation of summaries,'' in \emph{Proc. Text Summarization Branches Out}, Jul. 2004, pp. 74--81.

\bibitem{hamamci2024generatect}
I.~E. Hamamci, S.~Er, A.~Sekuboyina, E.~Simsar, A.~Tezcan, A.~G. Simsek, S.~N. Esirgun, F.~Almas, I.~Do{\u{g}}an, M.~F. Dasdelen \emph{et~al.}, ``Generatect: Text-conditional generation of 3d chest ct volumes,'' in \emph{Proc. Eur. Conf. Comput. Vis.}\hskip 1em plus 0.5em minus 0.4em\relax Springer, 2024, pp. 126--143.

\bibitem{kingma2014adam}
D.~P. Kingma and J.~Ba, ``Adam: A method for stochastic optimization,'' \emph{arXiv:1412.6980}, 2014.

\bibitem{dosovitskiy2021vit}
A.~Dosovitskiy, L.~Beyer, A.~Kolesnikov, D.~Weissenborn, X.~Zhai, T.~Unterthiner, M.~Dehghani, M.~Minderer, G.~Heigold, S.~Gelly, J.~Uszkoreit, and N.~Houlsby, ``An image is worth 16x16 words: Transformers for image recognition at scale,'' in \emph{Proc. Int. Conf. Learn. Represent.}, 2021.

\bibitem{draelos2021ctnet}
R.~L. Draelos, D.~Dov, M.~A. Mazurowski, J.~Y. Lo, R.~Henao, G.~D. Rubin, and L.~Carin, ``Machine-learning-based multiple abnormality prediction with large-scale chest computed tomography volumes,'' \emph{Med. Image Anal.}, vol.~67, p. 101857, 2021.

\bibitem{ronneberger2015unet}
O.~Ronneberger, P.~Fischer, and T.~Brox, ``U-net: Convolutional networks for biomedical image segmentation,'' in \emph{Proc. Int. Conf. Med. Image Comput. Comput.-Assisted Intervention}, 2015, pp. 234--241.

\bibitem{buzug2011computed}
T.~M. Buzug, ``Computed tomography,'' in \emph{Springer handbook of medical technology}.\hskip 1em plus 0.5em minus 0.4em\relax Springer, 2011, pp. 311--342.

\bibitem{remy1998multiplanar}
J.~Remy, M.~Remy-Jardin, D.~Artaud, and M.~Fribourg, ``Multiplanar and three-dimensional reconstruction techniques in ct: impact on chest diseases,'' \emph{Eur. Radiol.}, pp. 335--351, 1998.

\end{thebibliography}
\end{document}